\pdfoutput=1

\documentclass[11pt]{article}

\usepackage{EMNLP2023}

\usepackage{times}
\usepackage{latexsym}

\usepackage[T1]{fontenc}

\usepackage[utf8]{inputenc}

\usepackage{microtype}

\usepackage{inconsolata}

\usepackage{booktabs}
\usepackage{amsmath}
\usepackage{amssymb}
\usepackage{amsfonts}
\usepackage{pifont}
\newcommand{\cmark}{\ding{51}}%
\newcommand{\xmark}{\ding{55}}%
\usepackage{xcolor}
\usepackage{graphicx}
\usepackage{multirow}
\usepackage{algorithm}
\usepackage{algorithmicx}
\usepackage{algpseudocode}
\usepackage{listings}
\usepackage{pythonhighlight}

\newcommand{\suda}{$^{\heartsuit}$}
\newcommand{\hw}{$^{\clubsuit}$}

\title{Mirror: A Universal Framework for Various Information Extraction Tasks}
\author{
Tong Zhu\suda, Junfei Ren\suda, Zijian Yu\suda, Mengsong Wu\suda, Guoliang Zhang\suda, Xiaoye Qu\hw, \\\textbf{Wenliang Chen}\suda \thanks{\enspace Corresponding author}, \textbf{Zhefeng Wang}\hw, \textbf{Baoxing Huai}\hw, \textbf{Min Zhang}\suda \\
\suda~Institute of Artificial Intelligence, School of Computer Science and Technology, \\ Soochow University, China \\
\hw~Huawei Cloud, China \\
\texttt{\{tzhu7,jfrenjfren,zjyu,mswumsw,glzhang\}@stu.suda.edu.cn} \\
\texttt{\{quxiaoye,wangzhefeng,huaibaoxing\}@huawei.com} \\
\texttt{\{wlchen,minzhang\}@suda.edu.cn}
}

\begin{document}
\maketitle

\begin{abstract}

Sharing knowledge between information extraction tasks has always been a challenge due to the diverse data formats and task variations.
Meanwhile, this divergence leads to information waste and increases difficulties in building complex applications in real scenarios.
Recent studies often formulate IE tasks as a triplet extraction problem.
However, such a paradigm does not support multi-span and n-ary extraction, leading to weak versatility.
To this end, we reorganize IE problems into unified multi-slot tuples and propose a universal framework for various IE tasks, namely Mirror.
Specifically, we recast existing IE tasks as a multi-span cyclic graph extraction problem and devise a non-autoregressive graph decoding algorithm to extract all spans in a single step.
It is worth noting that this graph structure is incredibly versatile, and it supports not only complex IE tasks, but also machine reading comprehension and classification tasks.
We manually construct a corpus containing 57 datasets for model pretraining, and conduct experiments on 30 datasets across 8 downstream tasks.
The experimental results demonstrate that our model has decent compatibility and outperforms or reaches competitive performance with SOTA systems under few-shot and zero-shot settings.
The code, model weights, and pretraining corpus are available at \url{https://github.com/Spico197/Mirror} .

\end{abstract}
\section{Introduction}

Information Extraction (IE) is a fundamental field in Natural Language Processing (NLP), which aims to extract structured information from unstructured text~\cite{grishman_2019}, such as Named Entity Recognition (NER) \cite{qu2023distantly,gu2022delving,qu2023survey}, Relation Extraction (RE) \cite{cheng2021hacred}, Event Extraction (EE).
However, each IE task is usually isolated from specific data structures and delicate models, which makes it difficult to share knowledge across tasks~\cite{uie,genie}.

\begin{figure}[t]
    \centering
    \includegraphics[width=\columnwidth]{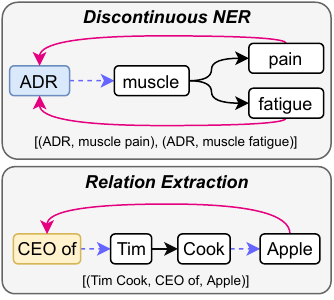}
    \caption{
        Multi-span cyclic graph for discontinuous NER and RE tasks (best viewed in color).
        The spans are connected by three types of edges, including \textbf{\textit{consecutive connections}}, dotted \textbf{\color[HTML]{695efb} \textit{jump connections}} and \textbf{\color[HTML]{E9087F} \textit{tail-to-head connections}}.
        \textit{ADR} in discontinuous NER refers to the entity label of Adverse Drug Reaction.
    }
    \label{fig:multi-span-cyclic-graph}
\end{figure}

\begin{table*}[t]
    \centering
        \begin{tabular}{c|ccccc|c}
        \toprule
        Model       & TANL    & UIE      & DeepStruct & InstructUIE & USM        & Mirror \\
        \midrule
        PLM         & T5-base & T5-large & GLM        & FlanT5 & RoBERTa    & DeBERTa-v3 \\
        \#Params    & 220M    & 770M     & 10B        & 11B & large 372M & large 434M \\
        \midrule
        Decoding    & AR                      & AR                      & AR                      & AR                      & NAR                          & NAR                          \\
        Indexing    & Partly & \textcolor{red}{\xmark} & \textcolor{red}{\xmark} & \textcolor{red}{\xmark} & {\color[HTML]{008114}\cmark} & {\color[HTML]{008114}\cmark} \\
        \midrule
        Triplet     & {\color[HTML]{008114}\cmark} & {\color[HTML]{008114}\cmark} & {\color[HTML]{008114}\cmark} & {\color[HTML]{008114}\cmark} & {\color[HTML]{008114}\cmark} & {\color[HTML]{008114}\cmark} \\
        Single-span NER & {\color[HTML]{008114}\cmark} & {\color[HTML]{008114}\cmark} & {\color[HTML]{008114}\cmark} & {\color[HTML]{008114}\cmark} & {\color[HTML]{008114}\cmark} & {\color[HTML]{008114}\cmark} \\
        Multi-span  & \textcolor{red}{\xmark} & $\circ$                 & $\circ$                 & $\circ$                 & \textcolor{red}{\xmark} & {\color[HTML]{008114}\cmark} \\
        N-ary tuple      & \textcolor{red}{\xmark} & \textcolor{red}{\xmark} & \textcolor{red}{\xmark} & \textcolor{red}{\xmark} & \textcolor{red}{\xmark} & {\color[HTML]{008114}\cmark} \\
        Classification        & \textcolor{red}{\xmark} & \textcolor{red}{\xmark} & \textcolor{red}{\xmark} & $\circ$                 & \textcolor{red}{\xmark} & {\color[HTML]{008114}\cmark} \\
        MRC         & \textcolor{red}{\xmark} & \textcolor{red}{\xmark} & \textcolor{red}{\xmark} & \textcolor{red}{\xmark} & $\circ$                 & {\color[HTML]{008114}\cmark} \\
        \bottomrule
        \end{tabular}
    \caption{
        Comparisons among systems.
        \textbf{Circle} $\circ$: the model may support the task theoretically, but the current implementation is not available.
        \textbf{AR}: the auto-regressive decoding while \textbf{NAR} is non-autoregressive.
        \textbf{Indexing}: whether the model could provide exact position information. TANL partly supports indexing because the generated tail entity in relation extraction is text-based without position information. Please refer to Appendix~\ref{sec:indexing} for more detailed comparisons.
        \textbf{Triplet}: ``(head, relation, tail)'' triplet extraction.
        \textbf{Single-span NER}: flat and nested NER tasks with consecutive spans.
        \textbf{Multi-span}: multi-span extraction, e.g., the discontinuous NER.
        \textbf{N-ary tuple}: the ability of n-ary tuple extraction, e.g., quadruple extraction.
        \textbf{Classification}: the classification tasks.
        \textbf{MRC}: extractive machine reading comprehension tasks.
        It is worth noting that generative models (TANL, UIE, DeepStruct, and InstructUIE) may be capable of all the tasks if their current paradigms or patterns are changed.
        However, since the original papers do not contain relevant experiments, we mark them as \textcolor{red}{\xmark} or $\circ$ here.
    }
    \label{tab:method_comparison}
\end{table*}

In order to unify the data formats and take advantage of common features between different tasks, there are two main routes in recent studies.
The first one is to utilize generative pretrained language models (PLMs) to generate the structured information directly.
\citet{uie} and \citet{tanl} structure the IE tasks as a sequence-to-sequence generation problem and use generative models to predict the structured information autoregressively.
However, such methods cannot provide the exact positions of the structured information, which is essential to the NER task and fair evaluations~\cite{devil-ee}.
Besides, the generation-based methods are usually slow and consume huge resources to train on large-scale datasets~\cite{deepstruct}.
The second way is to apply the extractive PLMs, which are faster to train and inference.
USM ~\cite{usm} regards the IE tasks as a triplet prediction problem via semantic matching.
However, this method is limited to a small range of triplet-based tasks, and it is unable to address multi-span and n-ary extraction problems.

To overcome the above challenges, we propose \textit{Mirror}, a novel framework that can handle complex multi-span extraction, n-ary extraction, machine reading comprehension (MRC), and even classification tasks, which are not supported by the previous universal IE systems.
As exemplified in Figure~\ref{fig:multi-span-cyclic-graph}, we formulate IE tasks as a unified multi-slot tuple extraction problem and transform those tuples into multi-span cyclic graphs.
This graph structure is rather flexible and scalable.
It can be applied to not only complex IE tasks but also MRC and classification tasks.
Mirror takes schemas as part of the model inputs, and this benefits few-shot and zero-shot tasks naturally.

Compared with other models in Table~\ref{tab:method_comparison}, Mirror supports efficient non-autoregressive decoding with position indexing and shows good compatibility across different tasks and datasets.
We conduct extensive experiments on 30 datasets from 8 tasks, including NER, RE, EE, Aspect-based Sentiment Analysis (ABSA), multi-span discontinuous NER, n-ary hyper RE, MRC, and classification.
To enhance the few-shot and zero-shot abilities, we manually collect 57 datasets across 5 tasks into a whole corpus for model pretraining.
The experimental results demonstrate that Mirror achieves competitive results under few-shot and zero-shot settings.

Our contributions are summarized as follows:
\begin{itemize}
    \item We propose a unified schema-guided multi-slot extraction paradigm, which is capable of complex information extraction, machine reading comprehension, and even classification tasks.
    \item We propose Mirror, a universal non-autoregressive framework that transforms multiple tasks into a multi-span cyclic graph.
    \item We conduct extensive experiments on 30 datasets from 8 tasks, and the results show that our model achieves competitive results under few-shot and zero-shot settings.
\end{itemize}

\section{Related Work}

\begin{figure*}[t]
    \centering
    \includegraphics[width=\textwidth]{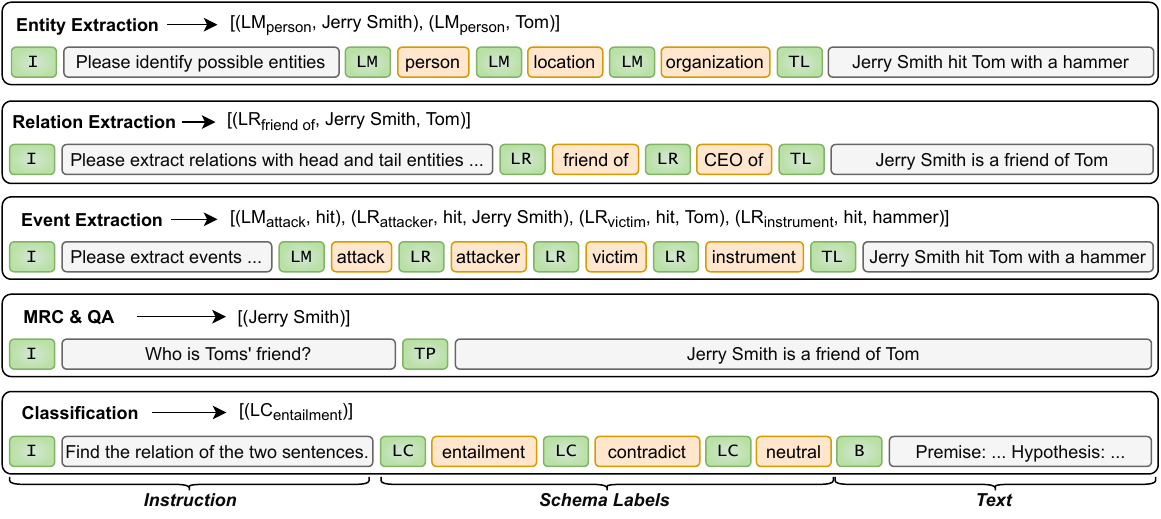}
    \caption{
        Unified data interface.
        We design a list of tokens to separate different parts:
        \texttt{[I]}: instruction.
        \texttt{[LM]}: mentions.
        \texttt{[LR]}: relations.
        \texttt{[LC]}: classifications.
        \texttt{[TL]}: text that connects with schema labels.
        \texttt{[TP]}: extractive MRC and QA texts without schema labels.
        \texttt{[B]}: the background text in the classification task.
    }
    \label{fig:unified-data-interface}
\end{figure*}

\subsection{Multi-task Information Extraction}

Multi-task IE has been a popular research topic in recent years.
The main idea is to use a single model to perform multiple IE tasks.
IE tasks could be formulated as different graph structures.
\citet{w2ner} formulate flat, nested, and discontinuous NER tasks as a graph with next-neighboring and tail-to-head connections.
Maximal cliques also have been used to flat \& discontinuous NER tasks \cite{mac-discontinuous-ner} and trigger-available \& trigger-free event extractions \cite{ptpcg}.
DyGIE++ takes NER, RE, and EE tasks as span graphs and applies iterative propagation to enhance spans' contextual representations \cite{dygiepp}.
OneIE uses a similar graph structure with global constraint features \cite{oneie}.

In addition to explicit graph-based multi-task IE systems, generative language models are widely used.
\citet{bart-ner} and \citet{bart-absa} add special index tokens into BART \cite{bart} vocabulary to help perform various NER and ABSA tasks and obtain explicit span positions.
TANL \cite{tanl} apply T5 \cite{t5} to generate texts with special enclosures as the predicted information.
GenIE \cite{genie} and DeepStruct \cite{deepstruct} share a similar idea to generate subject-relation-object triplets, and DeepStruct extends the model size to 10B with GLM \cite{glm}.

\subsection{Schema-guided Information Extraction}

In schema-guided IE systems, schemas are input as a guidance signal to help the model extract target information.
UIE \cite{uie} categorize IE tasks into span spotting and associating elementary tasks and devise a linearized query language.
\citet{lasuie} introduces the hyper relation extraction task to represent complex IE tasks like EE, and utilize external parsing tools to enhance the text representations.
InstructUIE \cite{instructuie} formulates schemas into instructions and uses FlanT5-11B \cite{flan-t5} to perform multi-task instruction tuning.

While the above methods utilize generative language models, they cannot predict exact positions, which brings ambiguity when evaluating~\cite{devil-ee}.
Besides, large generative language models are usually slow to train \& infer and require tons of computing resources.
USM \cite{usm} applies BERT-family models to extract triplets non-autoregressively.
USM regards IE as a unified schema matching task and uses a label-text matching model to extract triplets.
However, these methods cannot extend to complex IE tasks, such as multi-span discontinuous NER and n-ary information extractions.

\section{Mirror Framework}

\begin{figure*}[t]
    \centering
    \includegraphics[width=\textwidth]{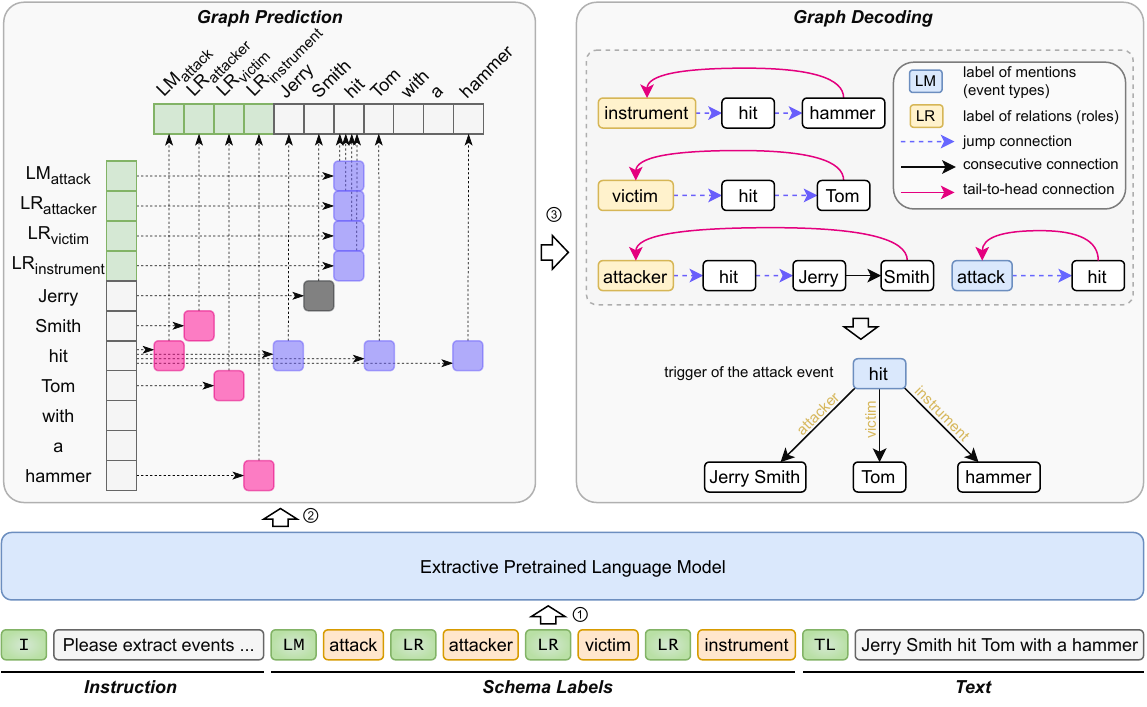}
    \caption{
        Model framework (best viewed in color).
        Mirror first constructs inputs for each task, then utilizes a pretrained language model to predict the adjacency matrix via the biaffine attention.
        After that, final results are decoded from the adjacency matrix accordingly.
    }
    \label{fig:model-framework}
\end{figure*}

In this section, we introduce the Mirror framework.
We first address the unified data input format to the model, then introduce the unified task formulation and the model structure.

\subsection{Unified Data Interface}

To enable the model to handle different IE tasks, we propose a unified data interface for the model input.
As shown in Figure~\ref{fig:unified-data-interface}, there are three parts: \textit{\textbf{instruction}}, \textit{\textbf{schema labels}}, and \textit{\textbf{text}}.
The \textbf{\textit{instruction}} is composed of a leading token \verb|[I]| and a natural language sentence.
The \verb|[I]| token indicates the instruction part while the sentence tells the model what it should do.
For example, the instruction of NER could be ``\textit{Please identify possible entities}''.
In MRC and Question Answering (QA) tasks, the instruction is the question to answer.

The \textbf{\textit{schema labels}} are task ontologies for schema-guided extraction.
This part consists of special token labels (\verb|[LM]|, \verb|[LR]|, and \verb|[LC]|) and corresponding label texts.
Among the special tokens, \verb|[LM]| denotes the label of mentions (or event types), \verb|[LR]| denotes the label of relations (or argument roles), and \verb|[LC]| denotes the label of classes.

The \textbf{\textit{text}} part is the input text that the model should extract information from.
It is composed of a leading token (\verb|[TL]|, \verb|[TP]| or \verb|[B]|) and a natural language sentence.
If the leading token is \verb|[TL]|, the model should link labels from schema labels to spans in the text.
While the \verb|[TP]| token indicates the target spans are only in the text, and the model should extract information from the text without schema labels.
In classification tasks, the model should not extract anything from the text part.
So we use a special leading token \verb|[B]| (background) to distinguish it from the extractive text.

With the above three parts, we can formulate extractive MRC, classification, and IE tasks into a unified data interface, and the model can be trained in a unified way even if the model is not based on generative language models.
For the robust model training, we manually collect 57 datasets from 5 tasks to make a corpus for model pretraining.
The data statistics for each IE task are listed in Table~\ref{tab:pretrain-dataset-statistics}.
To balance the number of examples in each task, we set a different maximum number of samples $N_{\max}$ for each task dataset.
If the number of instances in a dataset is less than $N_{\max}$, we keep the original dataset unchanged and do not perform oversampling.
For NER, RE, and EE tasks, we manually design a set of instructions and randomly pick one of them for each sample.
MRC datasets some classification datasets have inborn questions, so the numbers of instruction are much higher than the others.
For detailed statistics on each dataset, please refer to Appendix~\ref{sec:appendix_data}.

\begin{table}[t]
    \centering
    \resizebox{\columnwidth}{!}{
    \begin{tabular}{lrrrr}
        \toprule
        Task & \#Dataset & $N_{\max}$ & \#Instruction & \#Instance \\
        \midrule
        NER & 15 & 20,000 & 42 & 171,609 \\
        Cls$^{\clubsuit}$ & 27 & 5,000 & 54,070 & 134,758 \\
        RE & 9 & 20,000 & 9 & 123,876 \\
        MRC$^{\heartsuit}$ & 5 & 30,000 & 75,200 & 85,658 \\
        EE & 1 & All & 40 & 2,898 \\
        \midrule
        Total & 57 & - & - & 518,799 \\
        \bottomrule
    \end{tabular}
    }
    \caption{
        Pretraining dataset statistics.
        $^{\clubsuit}$ Classification tasks contain multi-choice MRC datasets.
        $^{\heartsuit}$ MRC stands for both extractive QA and extractive MRC datasets.
    }
    \label{tab:pretrain-dataset-statistics}
\end{table}

\subsection{Multi-slot Tuple and Multi-span Cyclic Graph}

We formulate IE tasks as a unified multi-slot tuple extraction problem.
As exemplified in Figure~\ref{fig:unified-data-interface}, in the RE task, the model is expected to extract a three-slot tuple: \texttt{(relation, head entity, tail entity)}.
Here, the tuple is \texttt{(LR$_{\text{friend of}}$, Jerry Smith, Tom)}.
The length of tuple slots could vary across tasks, so Mirror is able to solve n-ary extraction problems.

As shown in Figure~\ref{fig:multi-span-cyclic-graph} and the top right of Figure~\ref{fig:model-framework}, we formulate multi-slot tuples into a unified multi-span cyclic graph, and regard labels as the leading tokens in schema labels.
There are three types of connections in the graph: the \textbf{\textit{consecutive}} connection, the \textbf{\color[HTML]{695efb} \textit{jump}} connection, and the \textbf{\color[HTML]{E9087F}\textit{tail-to-head}} connection.
The \textbf{\textit{consecutive}} connection is adopted to \textbf{spans in the same entity}.
For an entity with multiple tokens, the consecutive connection connects from the first to the last.
As shown in Figure~\ref{fig:model-framework}, ``Jerry'' connects to ``Smith''.
If there is only one token in an entity, the consecutive connection is not used.
The \textbf{\color[HTML]{695efb} \textit{jump}} connection connects \textbf{different slots} in a tuple.
Schema labels and spans from texts are in different slots, so they are connected in jump connections.
For instance, the head and tail entities of a relation triplet are in different slots, so they are connected in jump connections.
The \textbf{\color[HTML]{E9087F}\textit{tail-to-head}} connection helps \textbf{locate the graph boundaries}.
It connects from the last token of the last slot to the first token of the first slot in a tuple.

In practice, we convert the answer of each slot into text positions.
For schema labels, we use the position of leading tags instead of literal strings.
For text spans, the position is a one-digit number if there is only one character, otherwise the start and end positions are listed.
For example, the 3-slot relation tuple \texttt{(LR$_{\text{friend of}}$, Jerry Smith, Tom)} will be converted into (9 {\color[HTML]{695efb} $\vdots$} 16 $\to$ 17 {\color[HTML]{695efb} $\vdots$} 22), where $\vdots$ denotes the jump connection, $\to$ stands for the consecutive connection, 
9 is the position of \texttt{LR$_{\text{friend of}}$}, 16 and 17 express \textit{Jerry Smith}, and 22 is the position of \textit{Tom}.
There is also a tail-to-head connection from 22 to 9.
The corresponding graph decoding algorithm is shown in Algorithm~\ref{alg:mcg-decoding}.
During inference, we first find the forward chain (9,16,17,22) and then verify the chain with the tail-to-head connection (22$\to$9).
After that, the multi-slot tuple is obtained with jump connections(9$\vdots$16) and (17$\vdots$22).

\begin{algorithm}[t]
    \renewcommand{\algorithmicrequire}{\textbf{Input:}}
    \renewcommand{\algorithmicensure}{\textbf{Output:}}
    \caption{\textsc{Multi-span Cyclic Graph Decoding}}
    \label{alg:mcg-decoding}
    \begin{algorithmic}[1]
        \Require Adjacency matrix $\mathcal{A}$
        \Ensure A set of multi-slot tuples $\mathcal{T}$
        \State $\mathcal{T} \gets \{\}$
        \State $\tilde{\mathcal{A}} \gets \mathcal{A}^{c} | \mathcal{A}^{j}$ \Comment{merge consecutive and jump connections}
        \State Find forward chains $\mathcal{C}$ from $\tilde{\mathcal{A}}$
        \For {$c \in \mathcal{C}$}
            \Comment{find legal paths with tail-to-head connections}
            \If{$c$ meets the need in $\mathcal{A}^{t}$}
                \State split $c$ into a tuple $t$ via $\mathcal{A}^j$
                \State $\mathcal{T} \gets \mathcal{T} \cup t$
            \EndIf
        \EndFor
        \State \Return $\mathcal{T}$
    \end{algorithmic}
\end{algorithm}

\subsection{Model Structure}

With the unified data interface and the multi-span cyclic graph, we propose a unified model structure for IE tasks.
For each token $x_i$ from the inputs, Mirror transforms it into a vector $h_i \in \mathbb{R}^{d_h}$ via a BERT-style extractive pretrained language model (PLM).
We use biaffine attention \cite{biaffine} to obtain the adjacency matrix $\mathcal{A}$ of the multi-span cyclic graph.
Mirror calculates the linking probability $p_{ij}^{k}, k \in \{\text{consecutive}, \text{jump}, \text{tail-to-head}\}$ between $x_i$ and $x_j$ as Equation~\ref{eqn:sim_calc} shows.
The final $\mathcal{A}$ is obtained via thresholding ($\mathcal{A}_{ij}^{k}=1$ if $p_{ij}^{k} > 0.5$ else 0).
\begin{equation}
    \label{eqn:sim_calc}
    \begin{gathered}
    \tilde{h_i} = \text{FFNN}_{s}\left(h_i\right), \quad \tilde{h_j} = \text{FFNN}_{e}\left(h_j\right) \\
    p_{ij}^{k} = \text{sigmoid}\left(\tilde{h_i}^{\top} U \tilde{h_j} / \sqrt{d_h}\right),
    \end{gathered} \\
\end{equation}
where $\tilde{h_i},\tilde{h_j} \in \mathbb{R}^{d_b}$.
$U \in \mathbb{R}^{d_b \times 3 \times d_b}$ is the trainable parameter, and 3 denotes consecutive, jump, and tail-to-head connections.
$\text{FFNN}$ is the feed-forward neural network with rotary positional embedding as introduced in \citet{roformer}.
The FFNN comprises a linear transformation, a GELU activation function~\cite{gelu}, and dropout~\cite{dropout}.

During training, we adopt the imbalance-class multi-label categorical cross entropy~\cite{global_pointer} as the loss function:

\begin{equation}
    \label{eqn:loss}
    \mathcal{L}(i,j) = \log \left(1 + \sum_{\Omega_{\text{neg}}} e^{p_{ij}^{k}}\right) + \log \left(1 + \sum_{\Omega_{\text{pos}}} e^{-p_{ij}^{k}}\right),
\end{equation}
where $\Omega_{\text{neg}}$ stands for negative samples ($\mathcal{A}_{ij}^{k}=0$), and $\Omega_{\text{pos}}$ denotes positive samples ($\mathcal{A}_{ij}^{k}=1$).

\begin{table*}[t]
  \centering
  \resizebox{\textwidth}{!}{%
  \begin{tabular}{c|c|ccccc|cccc}
    \toprule
    \multirow{2}{*}{Task} & \multirow{2}{*}{Datasets} & \multirow{2}{*}{TANL} & \multirow{2}{*}{UIE} & \multirow{2}{*}{DeepStruct} & \multirow{2}{*}{InstructUIE} & \multirow{2}{*}{USM} &
      Mirror &
      Mirror &
      Mirror &
      Mirror \\
    & & & & & & & \shortstack{w/ PT \\ w/ Inst.} & \shortstack{w/ PT \\ w/o Inst.} & \shortstack{w/o PT \\ w/ Inst.} & \shortstack{w/o PT \\ w/o Inst.} \\
    \midrule
    \multirow{3}{*}{NER}
     & ACE04   & -     & 86.89 & -     & - & 87.62 & 87.16 & 86.39 & \textbf{87.66} & 87.26 \\
     & ACE05   & 84.90 & 85.78 & 86.90 & 86.66 & \textbf{87.14} & 85.34 & 85.70 & 86.72 & 86.45 \\
     & CoNLL03 & 91.70 & 92.99 & 93.00 & 92.94 & \textbf{93.16} & 92.73 & 91.93 & 92.11 & 92.97 \\
      \midrule
      \multirow{4}{*}{RE}
     & ACE05   & 63.70 & 66.06 & 66.80 & - & 67.88 & 67.86 & 67.86 & 64.88 & \textbf{69.02} \\
     & CoNLL04 & 71.40 & 75.00 & 78.30 & 78.48 & \textbf{78.84} & 75.22 & 72.96 & 71.19 & 73.58 \\
     & NYT     & -     & 93.54 & 93.30 & 90.47 & 94.07 & 93.85 & \textbf{94.25} & 93.95 & 93.31 \\
     & SciERC  & -     & 36.53 & -     & 45.15 & 37.36 & 36.89 & 37.12 & 36.66 & \textbf{40.50} \\
      \midrule
      \multirow{4}{*}{EE} &
      ACE05-Tgg  & 68.40 & 73.36 & 69.80 & \textbf{77.13} & 72.41 & 74.44 & 73.05 & 72.66 & 73.38 \\
     & ACE05-Arg & 47.60 & 54.79 & 56.20 & \textbf{72.94} & 55.83 & 55.88 & 54.73 & 56.51 & 57.87 \\
     & CASIE-Tgg & -     & 69.33 & -     & 67.80 & 71.73 & 71.81 & 71.60 & \textbf{73.09} & 71.40 \\
     & CASIE-Arg & -     & 61.30 & -     & \textbf{63.53} & 63.26 & 61.27 & 61.04 & 60.44 & 58.87 \\
      \midrule
      \multirow{4}{*}{ABSA}
     & 14-res & - & 74.52 & - & - & \textbf{77.26} & 75.06 & 74.24 & 76.05 & 75.89 \\
     & 14-lap & - & 63.88 & - & - & \textbf{65.51} & 64.08 & 62.48 & 59.56 & 60.42 \\
     & 15-res & - & 67.15 & - & - & \textbf{69.86} & 66.40 & 63.61 & 60.26 & 67.41 \\
     & 16-res & - & 75.07 & - & - & \textbf{78.25} & 74.24 & 75.40 & 73.13 & 77.46 \\
     \midrule
     \multicolumn{2}{c|}{Avg.} & - & 71.75 & - & - & 73.35 & 72.15 & 71.49 & 70.99 & 72.39 \\
    \bottomrule
    \end{tabular}%
  }
  \caption{Results on 13 IE benchmarks (ACE-Tgg and ACE-Arg are in the same dataset with different evaluation metrics). PT is the abbreviation of pretraining, and Inst. denotes the task instruction.}
  \label{tab:main-results}
\end{table*}

\section{Experiments}

\subsection{Experiment Setup}

We utilize DeBERTa-v3-large~\cite{deberta} as the PLM.
The biaffine size $d_b$ is 512 with a dropout rate of 0.3.
The epoch number of pretraining is 3 with a learning rate of 2e-5.
Please refer to Appendix~\ref{sec:app-hyper-param} for detailed hyper-param settings.

Datasets are processed following \citet{uie} (13 IE datasets in Table~\ref{tab:main-results}, and 4 datasets in Table~\ref{tab:few_shot}), \citet{w2ner} (CADEC in Table~\ref{tab:multi-span-n-ary-results}), \citet{hyperred} (HyperRED in Table~\ref{tab:multi-span-n-ary-results}), \citet{usm} (7 zero-shot NER datasets in Table~\ref{tab:zero_shot_ner}), \citet{squad-v2} (SQuAD v2.0 in Table~\ref{tab:mrc-cls-results}), and \citet{glue} (7 GLUE datasets in Table~\ref{tab:mrc-cls-results}).
Data statistics and metrics are listed in Appendix~\ref{sec:appendix_data}.

\subsection{Baselines}

We compare Mirror with generation-based TANL~\cite{tanl}, DeepStruct~\cite{deepstruct}, UIE~\cite{uie}, InstructUIE~\cite{instructuie}, and extraction-based USM~\cite{usm} in triplet-based IE tasks.
In the multi-span discontinuous NER task, we compare Mirror with task-specific BART-NER~\cite{bart-ner} and W2NER~\cite{w2ner}.
The baseline system in hyper RE is CubeRE~\cite{hyperred}.
As to MRC tasks, the baseline models are BERT~\cite{bert}, RoBERTa~\cite{roberta}, and DeBERTa-v3~\cite{deberta}.

\subsection{Main Results}

Mirror performances on 13 IE benchmarks are presented in Table~\ref{tab:main-results}.
Compared with other baseline models, Mirror surpasses baseline models on some datasets in NER (ACE04), RE (ACE05, NYT), and EE (CASIE-Trigger) tasks. 
When compared to extraction-based USM, Mirror achieves competitive results on most of tasks, while lagging in NER (ACE05), RE (CoNLL04), and EE (CASIE-Arg).
Compared to generation-based methods, Mirror outperforms TANL across all datasets and surpasses UIE in most datasets.
When the model parameter comes to 10B, DeepStruct outperforms Mirror on CoNLL04 in the RE task, while Mirror reaches very close results or outperforms DeepStruct on the other datasets.
InstructUIE (11B) demonstrates similar performance on NER datasets, while achieving high scores in RE (SciERC) and EE (ACE05-Tgg \& Arg), surpassing other models by a significant margin.
Apart from these datasets, InstructUIE performs about the same as UIE, USM, and Mirror.

We provide ablation studies on Mirror with different pretraining and fine-tuning strategies.
Performance degrades if either pretraining or instruction fine-tuning is not performed.
Mirror benefits from pretraining when utilizing instructions (w/ Inst.) and increases 0.66\% scores on average.
However, when instructions are discarded (w/o Inst.), pretraining (w/ PT) does not bring performance gain.
Pretraining has been confirmed on UIE and USM to enhance model performances, and it is crucial to enable the zero-shot inference ability.
However, based on the results from Table~\ref{tab:main-results}, we find that if Mirror is applied in one specific task with sufficient training resources, it may not need to perform the pretraining step (e.g., NYT dataset).

Besides the traditional IE tasks in Table~\ref{tab:main-results}, Mirror also supports multi-span discontinuous NER and n-ary hyper relation extraction as shown in Table~\ref{tab:multi-span-n-ary-results}.
We provide Mirror (w/ PT, w/ Inst) and Mirror (w/o PT, w/o Inst.) results on CADEC according to their good performances on the IE tasks in Table~\ref{tab:main-results}.
However, Mirror is less powerful than task-specific SOTA models.
On the n-ary hyper relation extraction task, Mirror outperforms the task-specific model CubeRE and achieves new SOTA results.
Table~\ref{tab:multi-span-n-ary-results} indicates Mirror's compatibility with complex multi-span and n-ary extraction problems.

The above facts indicate that Mirror has good compatibility across different IE problems, and we extend the universal IE system to complex multi-span and n-ary extraction tasks, which are not supported by previous universal IE systems.

\begin{table}[t]
  \centering
  \begin{tabular}{lrrr}
    \toprule
    & \multicolumn{1}{c}{P} & \multicolumn{1}{c}{R} & \multicolumn{1}{c}{F1} \\
    \midrule
    \multicolumn{4}{l}{\textit{Discontinuous NER: CADEC}} \\
    BART-NER & 70.08 & 71.21 & 70.64 \\
    W2NER    & 74.09 & \textbf{72.35} & \textbf{73.21} \\
    Mirror$_{\text{w/ PT \& Inst.}}$   & \textbf{74.83} & 65.45 & 69.83 \\
    Mirror$_{\text{w/o PT \& Inst.}}$ & 68.80 & 68.38 & 68.59 \\
    \midrule
    \multicolumn{4}{l}{\textit{N-ary Tuples: HyperRED}} \\
    CubeRE   & 66.39 & \textbf{67.12} & 66.75 \\
    Mirror$_{\text{w/ PT \& Inst.}}$   & 71.29 & 62.46 & 66.58  \\
    Mirror$_{\text{w/o PT \& Inst.}}$ & \textbf{75.41} & 61.14 & \textbf{67.53}  \\
    \bottomrule
  \end{tabular}%
  \caption{
    Results on multi-span and n-ary information extraction tasks.
    Tgg and Arg in Event Extraction refer to Trigger (Event Detection) and Argument (Event Argument Extraction), respectively.
  }
  \label{tab:multi-span-n-ary-results}
\end{table}

\subsection{Few-shot Results}

Followed by \citet{uie} and \citet{usm}, we analyze Mirror's few-shot ability on NER, RE, EE, and ABSA tasks.
As shown in Table~\ref{tab:few_shot}, Mirror (w/ PT, w/ Inst.) outperforms USM and achieves SOTA results on CoNLL03, ACE05, and 16-res datasets.
In the RE task on CoNLL04, the best model is USM, achieving an average score of 50.12, while Mirror is less effective with only 43.16 average scores.
Among all the four tasks, NER may be relatively easier for the model to deal with.
The 10-shot NER score of Mirror is 84.69, while the fine-tuned Mirror on the full dataset gets an F1 score of 92.73.
The gaps on other datasets between 10-shot and fully fine-tuned results are larger, indicating the task difficulties.

\begin{table}[t]
    \centering
    \resizebox{\columnwidth}{!}{%
    \begin{tabular}{c|c|ccc|c}
    \toprule
    Task & Model  & 1-shot & 5-shot & 10-shot & Avg.  \\
    \midrule
    \multirow{3}{*}{\shortstack{NER\\CoNLL03}}
     & UIE    & 57.53  & 75.32  & 79.12   & 70.66 \\
     & USM    & 71.11  & \textbf{83.25}  & 84.58   & 79.65 \\
     & Mirror & \textbf{76.49} & 82.45 & \textbf{84.69} & \textbf{81.21} \\
    \midrule
    \multirow{3}{*}{\shortstack{RE\\CoNLL04}}
    & UIE    & 34.88  & 51.64  & 58.98   & 48.50 \\
    & USM    & \textbf{36.17}  & \textbf{53.20}  & \textbf{60.99}   & \textbf{50.12} \\
    & Mirror &     26.29   &   47.42     &   55.77      &  43.16     \\
    \midrule
    \multirow{3}{*}{\shortstack{Event Trigger\\ACE05}}
    & UIE    & 42.37  & 53.07  & 54.35   & 49.93 \\
    & USM    & 40.86  & 55.61  & 58.79   & 51.75 \\
    & Mirror &   \textbf{47.77}     & \textbf{57.90}       &  \textbf{59.16}       & \textbf{54.94}      \\
    \midrule
    \multirow{3}{*}{\shortstack{Event Arg\\ACE05}}
    & UIE    & 14.56  & 31.20  & 35.19   & 26.98 \\
    & USM    & 19.01  & 36.69  & \textbf{42.48}   & 32.73 \\
    & Mirror &    \textbf{23.18}    & \textbf{37.74}       & 39.20        & \textbf{33.38}      \\
    \midrule
    \multirow{3}{*}{\shortstack{ABSA\\16-res}}
    & UIE    & 23.04  & 42.67  & 53.28   & 39.66 \\
    & USM    & 30.81  & \textbf{52.06}  & 58.29   & 47.05 \\
    & Mirror &      \textbf{36.21}  & 51.65       & \textbf{58.59}        & \textbf{48.82} \\
    \bottomrule
    \end{tabular}%
    }
    \caption{
        Few-shot results on IE tasks.
        These datasets are not included in the pretraining phase of Mirror.
    }
    \label{tab:few_shot}
\end{table}

\subsection{Zero-shot Results}

Table~\ref{tab:zero_shot_ner} shows the zero-shot performances on 7 NER datasets.
These datasets are not included in pretraining, and we use the pretrained Mirror to make predictions directly.
The results show that Mirror outperforms USM by a large margin (an average F1 score of 9.44), and it is very competitive with InstructUIE (FlanT5-11B).
Considering the model scale, Mirror is surprisingly good at zero-shot NER tasks.
However, ChatGPT is very powerful in the zero-shot NER task and achieves absolute SOTA performance.
Except for simple model scaling, we may need to collect a more diverse pretraining corpus for better results.

\begin{table*}[t]
    \centering
    \begin{tabular}{l|cccccccc}
        \toprule
    Model &
      \multicolumn{1}{c}{Movie} &
      \multicolumn{1}{c}{Restaurant} &
      \multicolumn{1}{c}{AI} &
      \multicolumn{1}{c}{Literature} &
      \multicolumn{1}{c}{Music} &
      \multicolumn{1}{c}{Politics} &
      \multicolumn{1}{c}{Science} &
      \multicolumn{1}{c}{Avg.} \\
      \midrule
    Davinci          & 0.84           & 2.94           & 2.97           & 9.87           & 13.83          & 18.42          & 10.04          & 8.42           \\
    ChatGPT          &  41.00    & 37.76 & 54.40 &  54.07    & 61.24 &  59.12    & 63.00 & 52.94 \\
    \midrule
    USM              & 37.73          & 14.73          & 28.18          & \textbf{56.00} & 44.93          & 36.10          & 44.09          & 37.39          \\
    InstructUIE      & \textbf{63.00} &  \textbf{20.99}    & \textbf{49.00}          & 47.21          & 53.61          & 48.15          & 49.30          & \textbf{47.32}          \\
    Mirror$_{\text{direct}}$           & 39.20 & 16.32 & 45.23 & 46.32 & \textbf{58.61} & \textbf{67.30} & \textbf{54.84} & 46.83  \\
    \bottomrule
    \end{tabular}%
    \caption{
        Zero-shot results on 7 NER datasets.
        Results of Davinci and ChatGPT are derived from \citet{instructuie}.
        Mirror$_{\text{direct}}$ is the pretrained Mirror w/ Inst. while these datasets are not included in the pretraining phase.
    }
    \label{tab:zero_shot_ner}
\end{table*}

\begin{table*}[t]
  \centering

  \begin{tabular}{l|c|ccccccc}
    \toprule
    \multirow{2}{*}{Model} & SQuAD 2.0 & CoLA & QQP & MNLI & SST-2 & QNLI & RTE & MRPC \\
    & (EM/F1) & (Mcc) & (Acc) & (Acc) & (Acc) & (Acc) & (Acc) & (Acc) \\
    \midrule
    BERT-large         & 79.0 / 81.8         & 60.6       & 91.3      & -          & 93.2        & 92.3       & 70.4      & 84.1       \\
    RoBERTa-large      & 86.5 / 89.4         & 68.0       & 92.2      & 90.2       & 96.4        & 93.9       & 86.6      & 88.8       \\
    DeBERTa v3-large   & 89.0 / 91.5         & 75.3       & 93.0      & 91.9       & 96.9        & 96.0       & 92.7      & 92.2       \\
    Mirror$_{\text{direct}}$ & 40.4 / 67.4         & 63.9       & 84.8      & 85.9       & 93.6        & 91.6       & 85.9      & 89.2 \\
    \bottomrule
  \end{tabular}
  \caption{
    Results on MRC and classification tasks.
    We list Mirror performance on SQuAD 2.0 development set and GLUE development sets.
    Baseline results are derived from \citet{deberta}.
    Because SQuAD v2 and GLUE datasets are included in Mirror pretraining for 3 epochs, we direct make inferences with the pretrained model (noted as Mirror$_\text{direct}$, the same model used in zero-shot NER), and do not perform further fine-tuning, while other baselines are fine-tuned with a full dataset on every single task.
  }
  \label{tab:mrc-cls-results}
\end{table*}

\subsection{Results on MRC and Classification}

To show the model compatibility on extractive MRC and classification tasks, we conduct experiments on SQuAD v2 and GLUE language understanding benchmarks.
The experimental results are demonstrated in Table~\ref{tab:mrc-cls-results}.
Comparing the results in \citet{deberta}, we do not report performance on the STS-B dataset since Mirror's extraction paradigm does not support the regression task.
Although Mirror$_{\text{direct}}$ does not perform full fine-tuning like the other systems, it still produces competitive results.
It outperforms BERT-large on CoLA and SST-2 and is better than RoBERTa-large on MRPC.
The results indicate that Mirror is capable of various tasks besides information extraction.
We leave full fine-tuning for future work to improve Mirror performances.

\subsection{Analysis on Label Span Types}

Mirror adopts the leading token in schema labels (\verb|[LC]|, \verb|[LM]| and \verb|[LR]|) as the label span that connects to target text spans.
To analyze the effect of different label span types, we conduct experiments to change the leading token into a literal content string.
In other words, in a NER task that extract \textit{person} entities, we compare the effect of \verb|[LM]| token and \verb|person| string as the label span.

The results are demonstrated in Table~\ref{tab:label_span}.
We find that the label type does not bring too many differences.
In Mirror w/ Inst., the literal content string is slightly better than bare tags with only a 0.19 F1 score advantage.
While in Mirror w/o Inst., the tag-based method surpasses the content-based method by 0.72 F1 scores.
Similar to \citet{mtb}, these results show that although the label tag is a simple token without pretraining, it does not affect the model's ability to incorporate features from global and local contexts.

\begin{table}[t]
    \centering
    \resizebox{\columnwidth}{!}{
        \begin{tabular}{c|c|ccc}
        \toprule
        w/ Inst. & Label Type & P & R & F1  \\
        \midrule
        \multirow{2}{*}{{\color[HTML]{008114}\cmark}} & Tag & 92.12  & 92.10  & 92.11   \\
         & Content & 91.89  & 92.71  & 92.30   \\
         \midrule
        \multirow{2}{*}{\textcolor{red}{\xmark}} & Tag & 92.79  & 93.15  & 92.97   \\
         & Content & 91.91  & 92.58  & 92.25  \\
        \bottomrule
        \end{tabular}
    }
    \caption{
        Results on different label span types.
        This experiment is conducted on the CoNLL03 dataset w/o pretraining.
    }
    \label{tab:label_span}
\end{table}

\subsection{Analysis on Pretraining Datasets}

\begin{table*}[t]
    \centering
    \begin{tabular}{lccccc}
    \toprule
     Pretrain Data &  \shortstack{NER (F1) \\ CoNLL03}&  \shortstack{RE (F1) \\ NYT } &  \shortstack{MRC (F1) \\ SQuAD v2}& \shortstack{Cls (Acc) \\ MRPC } & Average\\
     \midrule
     All&  66.91&  69.67&  67.39&  89.22& 73.30\\
     \quad w/o Cls&  66.82&  57.03&  67.14&  0.00& 47.75\\
     \quad w/o IE&  0.00&  0.00&  68.77&  89.22& 39.50\\
     \quad w/o Span&  66.76&  54.76&  0.00&  87.50& 52.26\\
     \bottomrule
    \end{tabular}
    \caption{Ablation study on the pretraining data. We evaluate the pretrained Mirror$_{\text{direct}}$ without further fine-tuning.}
    \label{tab:ablation_pretrain_data}
\end{table*}

Traditionally, the classification task is different from the extraction task as they optimize different objectives.
Since Mirror unifies the two tasks into one framework, it is interesting to find how they affect each other in the pretraining phase.
We provide an ablation study on different types of pretraining data in Table~\ref{tab:ablation_pretrain_data}.
It is surprising that pretraining on classification datasets help improve the extraction tasks, and relation extraction is the most affected one.
This may be due to the similarity between relation labels and semantic class labels.
It is also interesting that span-based datasets (e.g. MRC datasets) are beneficial to the classification task (87.50 → 89.22).
Overall, all kinds of the pretraining datasets bring greater mutual benefits and improve the model performance.

\subsection{Analysis on Inference Speed}

We conduct speed tests on the CoNLL03's validation set with one NVIDIA V100 GPU under the same environment.
The results are presented in Table~\ref{tab:inference_speed}.
Compared to the popular generative T5-large UIE model~\cite{uie}, our model is up to 32.61 times faster when inference, and the advantage grows when increasing the batch size from 1 to 2.

\begin{table}[h]
    \centering
    \begin{tabular}{crrr}
    \toprule
         batch size&  UIE &  Mirror & Speed-Up
\\
    \midrule
         1&  0.21&  5.68& 27.24
\\
 2& 0.32& 10.56&32.61
\\
    \bottomrule
    \end{tabular}
    \caption{Inference speed (instances per second) test on CoNLL03 validation set.}
    \label{tab:inference_speed}
\end{table}

\section{Conclusion}

We propose Mirror, a schema-guided framework for universal information extraction.
Mirror transforms IE tasks into a unified multi-slot tuple extraction problem and introduces the multi-span cyclic graph to represent such structures.
Due to the flexible design, Mirror is capable of multi-span and n-ary extraction tasks.
Compared to previous systems, Mirror supports not only complex information extraction but also MRC and classification tasks.
We manually collect 57 datasets for pretraining and conduct experiments on 30 datasets across 8 tasks.
The experimental results show good compatibility, and Mirror achieves competitive performances with state-of-the-art systems.

\section*{Limitations}

Content input length: Due to the backbone DeBERTa model constraint, the maximal sequence length is 512 and can hardly extend to longer texts.
This limits the exploration of tasks with many schema labels and document-level IE.

Multi-turn result modification: Mirror predicts the multi-span cyclic graph in a paralleled non-autoregressive style.
Although it is efficient in training and inference, it may lack global history knowledge from previous answers.

Data format unification: There are many IE tasks, and the formats may vary a lot.
Although the current unified data interface supports most common tasks, it may not be practical for some tasks.

Lack of large-scale event datasets for pretraining: There are many NER and RE datasets.
However, there are few large-scale event extraction corpus with high diversity in domains and schemas, which may limit the model performance on event-relevant information extraction tasks.

\section*{Acknowledgments}

This work is supported by the National Natural Science Foundation of China (Grant No. 61936010) and Provincial Key Laboratory for Computer Information Processing Technology, Soochow University.
This work is also supported by Collaborative Innovation Center of Novel Software Technology and Industrialization, the Priority Academic Program Development of Jiangsu Higher Education Institutions, and the joint research project of Huawei Cloud and Soochow University.
We would also like to thank the anonymous reviewers for their insightful and valuable comments.

\bibliography{anthology,custom}

\begin{thebibliography}{102}
\expandafter\ifx\csname natexlab\endcsname\relax\def\natexlab#1{#1}\fi

\bibitem[{Baldini~Soares et~al.(2019)Baldini~Soares, FitzGerald, Ling, and Kwiatkowski}]{mtb}
Livio Baldini~Soares, Nicholas FitzGerald, Jeffrey Ling, and Tom Kwiatkowski. 2019.
\newblock \href {https://doi.org/10.18653/v1/P19-1279} {Matching the blanks: Distributional similarity for relation learning}.
\newblock In \emph{Proceedings of the 57th Annual Meeting of the Association for Computational Linguistics}, pages 2895--2905, Florence, Italy. Association for Computational Linguistics.

\bibitem[{Bjerva et~al.(2020)Bjerva, Bhutani, Golshan, Tan, and Augenstein}]{SubjQA}
Johannes Bjerva, Nikita Bhutani, Behzad Golshan, Wang-Chiew Tan, and Isabelle Augenstein. 2020.
\newblock \href {https://doi.org/10.18653/v1/2020.emnlp-main.442} {{SubjQA}: {A} {D}ataset for {S}ubjectivity and {R}eview {C}omprehension}.
\newblock In \emph{Proceedings of the 2020 Conference on Empirical Methods in Natural Language Processing (EMNLP)}, pages 5480--5494, Online. Association for Computational Linguistics.

\bibitem[{Bowman et~al.(2015)Bowman, Angeli, Potts, and Manning}]{SNLI}
Samuel~R. Bowman, Gabor Angeli, Christopher Potts, and Christopher~D. Manning. 2015.
\newblock \href {https://doi.org/10.18653/v1/d15-1075} {A large annotated corpus for learning natural language inference}.
\newblock In \emph{Proceedings of the 2015 Conference on Empirical Methods in Natural Language Processing, {EMNLP} 2015, Lisbon, Portugal, September 17-21, 2015}, pages 632--642. The Association for Computational Linguistics.

\bibitem[{Chen and Li(2021)}]{Wiki-ZSL}
Chih-Yao Chen and Cheng-Te Li. 2021.
\newblock \href {http://arxiv.org/abs/2104.04697} {Zs-bert: Towards zero-shot relation extraction with attribute representation learning}.

\bibitem[{Chen et~al.(2022)Chen, Xu, Zhang, and Huang}]{HarveyNER}
Pei Chen, Haotian Xu, Cheng Zhang, and Ruihong Huang. 2022.
\newblock \href {https://doi.org/10.18653/v1/2022.naacl-main.243} {Crossroads, buildings and neighborhoods: A dataset for fine-grained location recognition}.
\newblock In \emph{Proceedings of the 2022 Conference of the North American Chapter of the Association for Computational Linguistics: Human Language Technologies}, pages 3329--3339, Seattle, United States. Association for Computational Linguistics.

\bibitem[{Cheng et~al.(2021)Cheng, Liu, Qu, Zhao, Liang, Wang, Huai, Yuan, and Xiao}]{cheng2021hacred}
Qiao Cheng, Juntao Liu, Xiaoye Qu, Jin Zhao, Jiaqing Liang, Zhefeng Wang, Baoxing Huai, Nicholas~Jing Yuan, and Yanghua Xiao. 2021.
\newblock Hacred: A large-scale relation extraction dataset toward hard cases in practical applications.
\newblock In \emph{Findings of the Association for Computational Linguistics: ACL-IJCNLP 2021}, pages 2819--2831.

\bibitem[{Chia et~al.(2022)Chia, Bing, Aljunied, Si, and Poria}]{hyperred}
Yew~Ken Chia, Lidong Bing, Sharifah~Mahani Aljunied, Luo Si, and Soujanya Poria. 2022.
\newblock \href {https://aclanthology.org/2022.emnlp-main.688} {A dataset for hyper-relational extraction and a cube-filling approach}.
\newblock In \emph{Proceedings of the 2022 Conference on Empirical Methods in Natural Language Processing}, pages 10114--10133, Abu Dhabi, United Arab Emirates. Association for Computational Linguistics.

\bibitem[{Chung et~al.(2022)Chung, Hou, Longpre, Zoph, Tay, Fedus, Li, Wang, Dehghani, Brahma, Webson, Gu, Dai, Suzgun, Chen, Chowdhery, Narang, Mishra, Yu, Zhao, Huang, Dai, Yu, Petrov, Chi, Dean, Devlin, Roberts, Zhou, Le, and Wei}]{flan-t5}
Hyung~Won Chung, Le~Hou, Shayne Longpre, Barret Zoph, Yi~Tay, William Fedus, Eric Li, Xuezhi Wang, Mostafa Dehghani, Siddhartha Brahma, Albert Webson, Shixiang~Shane Gu, Zhuyun Dai, Mirac Suzgun, Xinyun Chen, Aakanksha Chowdhery, Sharan Narang, Gaurav Mishra, Adams Yu, Vincent~Y. Zhao, Yanping Huang, Andrew~M. Dai, Hongkun Yu, Slav Petrov, Ed~H. Chi, Jeff Dean, Jacob Devlin, Adam Roberts, Denny Zhou, Quoc~V. Le, and Jason Wei. 2022.
\newblock \href {https://doi.org/10.48550/arXiv.2210.11416} {Scaling instruction-finetuned language models}.
\newblock \emph{CoRR}, abs/2210.11416.

\bibitem[{Clark et~al.(2018)Clark, Cowhey, Etzioni, Khot, Sabharwal, Schoenick, and Tafjord}]{ARC}
Peter Clark, Isaac Cowhey, Oren Etzioni, Tushar Khot, Ashish Sabharwal, Carissa Schoenick, and Oyvind Tafjord. 2018.
\newblock Think you have solved question answering? try arc, the {AI2} reasoning challenge.
\newblock \emph{CoRR}, abs/1803.05457.

\bibitem[{Derczynski et~al.(2016)Derczynski, Bontcheva, and Roberts}]{Broad_Twitter_Corpus}
Leon Derczynski, Kalina Bontcheva, and Ian Roberts. 2016.
\newblock \href {https://aclanthology.org/C16-1111} {Broad {T}witter corpus: A diverse named entity recognition resource}.
\newblock In \emph{Proceedings of {COLING} 2016, the 26th International Conference on Computational Linguistics: Technical Papers}, pages 1169--1179, Osaka, Japan. The COLING 2016 Organizing Committee.

\bibitem[{Devlin et~al.(2019)Devlin, Chang, Lee, and Toutanova}]{bert}
Jacob Devlin, Ming-Wei Chang, Kenton Lee, and Kristina Toutanova. 2019.
\newblock \href {https://doi.org/10.18653/v1/N19-1423} {{BERT}: Pre-training of deep bidirectional transformers for language understanding}.
\newblock In \emph{Proceedings of the 2019 Conference of the North {A}merican Chapter of the Association for Computational Linguistics: Human Language Technologies, Volume 1 (Long and Short Papers)}, pages 4171--4186, Minneapolis, Minnesota. Association for Computational Linguistics.

\bibitem[{Dogan et~al.(2014)Dogan, Leaman, and Lu}]{NCBIDisease}
Rezarta~Islamaj Dogan, Robert Leaman, and Zhiyong Lu. 2014.
\newblock Ncbi disease corpus: A resource for disease name recognition and concept normalization.
\newblock \emph{Journal of biomedical informatics}, 47:1--10.

\bibitem[{Dolan and Brockett(2005)}]{mrpc}
William~B. Dolan and Chris Brockett. 2005.
\newblock \href {https://aclanthology.org/I05-5002/} {Automatically constructing a corpus of sentential paraphrases}.
\newblock In \emph{Proceedings of the Third International Workshop on Paraphrasing, IWP@IJCNLP 2005, Jeju Island, Korea, October 2005, 2005}. Asian Federation of Natural Language Processing.

\bibitem[{Dozat and Manning(2017)}]{biaffine}
Timothy Dozat and Christopher~D. Manning. 2017.
\newblock \href {https://openreview.net/forum?id=Hk95PK9le} {Deep biaffine attention for neural dependency parsing}.
\newblock In \emph{5th International Conference on Learning Representations, {ICLR} 2017, Toulon, France, April 24-26, 2017, Conference Track Proceedings}. OpenReview.net.

\bibitem[{Du et~al.(2022)Du, Qian, Liu, Ding, Qiu, Yang, and Tang}]{glm}
Zhengxiao Du, Yujie Qian, Xiao Liu, Ming Ding, Jiezhong Qiu, Zhilin Yang, and Jie Tang. 2022.
\newblock \href {https://doi.org/10.18653/v1/2022.acl-long.26} {{GLM}: General language model pretraining with autoregressive blank infilling}.
\newblock In \emph{Proceedings of the 60th Annual Meeting of the Association for Computational Linguistics (Volume 1: Long Papers)}, pages 320--335, Dublin, Ireland. Association for Computational Linguistics.

\bibitem[{Fei et~al.(2022)Fei, Wu, Li, Li, Li, Qin, Zhang, Zhang, and Chua}]{lasuie}
Hao Fei, Shengqiong Wu, Jingye Li, Bobo Li, Fei Li, Libo Qin, Meishan Zhang, Min Zhang, and Tat{-}Seng Chua. 2022.
\newblock \href {http://papers.nips.cc/paper\_files/paper/2022/hash/63943ee9fe347f3d95892cf87d9a42e6-Abstract-Conference.html} {Lasuie: Unifying information extraction with latent adaptive structure-aware generative language model}.
\newblock In \emph{NeurIPS}.

\bibitem[{Gardent et~al.(2017)Gardent, Shimorina, Narayan, and Perez-Beltrachini}]{WebNLG}
Claire Gardent, Anastasia Shimorina, Shashi Narayan, and Laura Perez-Beltrachini. 2017.
\newblock \href {https://doi.org/10.18653/v1/P17-1017} {Creating training corpora for {NLG} micro-planners}.
\newblock In \emph{Proceedings of the 55th Annual Meeting of the Association for Computational Linguistics (Volume 1: Long Papers)}, pages 179--188, Vancouver, Canada. Association for Computational Linguistics.

\bibitem[{Grishman(2019)}]{grishman_2019}
Ralph Grishman. 2019.
\newblock \href {https://doi.org/10.1017/S1351324919000512} {Twenty-five years of information extraction}.
\newblock \emph{Natural Language Engineering}, 25(6):677–692.

\bibitem[{Gu et~al.(2022)Gu, Qu, Wang, Zheng, Huai, and Yuan}]{gu2022delving}
Yingjie Gu, Xiaoye Qu, Zhefeng Wang, Yi~Zheng, Baoxing Huai, and Nicholas~Jing Yuan. 2022.
\newblock Delving deep into regularity: a simple but effective method for chinese named entity recognition.
\newblock \emph{arXiv preprint arXiv:2204.05544}.

\bibitem[{Guan et~al.(2023)Guan, Man, Chen, Yao, Hu, Zhu, Smith, Lim, and Yue}]{FindVehicle}
Runwei Guan, Ka~Lok Man, Feifan Chen, Shanliang Yao, Rongsheng Hu, Xiaohui Zhu, Jeremy~S. Smith, Eng~Gee Lim, and Yutao Yue. 2023.
\newblock \href {https://api.semanticscholar.org/CorpusID:258291946} {Findvehicle and vehiclefinder: A ner dataset for natural language-based vehicle retrieval and a keyword-based cross-modal vehicle retrieval system}.
\newblock \emph{ArXiv}, abs/2304.10893.

\bibitem[{Gurulingappa et~al.(2012)Gurulingappa, Rajput, Roberts, Fluck, Hofmann-Apitius, and Toldo}]{ADE}
Harsha Gurulingappa, Abdul~Mateen Rajput, Angus Roberts, Juliane Fluck, Martin Hofmann-Apitius, and Luca Toldo. 2012.
\newblock \href {https://doi.org/https://doi.org/10.1016/j.jbi.2012.04.008} {Development of a benchmark corpus to support the automatic extraction of drug-related adverse effects from medical case reports}.
\newblock \emph{Journal of Biomedical Informatics}, 45(5):885--892.
\newblock Text Mining and Natural Language Processing in Pharmacogenomics.

\bibitem[{Han et~al.(2018)Han, Zhu, Yu, Wang, Yao, Liu, and Sun}]{FewRel}
Xu~Han, Hao Zhu, Pengfei Yu, Ziyun Wang, Yuan Yao, Zhiyuan Liu, and Maosong Sun. 2018.
\newblock \href {http://arxiv.org/abs/1810.10147} {Fewrel: A large-scale supervised few-shot relation classification dataset with state-of-the-art evaluation}.

\bibitem[{Hao et~al.(2023)Hao, Xiaozhi, Feng, Kaisheng, Lei, Juanzi, Zhiyuan, and Weixing}]{devil-ee}
Peng Hao, Wang Xiaozhi, Yao Feng, Zeng Kaisheng, Hou Lei, Li~Juanzi, Liu Zhiyuan, and Shen Weixing. 2023.
\newblock \href {http://arxiv.org/abs/2306.06918} {The {Devil} is in the {Details}: {On} the {Pitfalls} of {Event} {Extraction} {Evaluation}}.
\newblock ArXiv:2306.06918 [cs].

\bibitem[{He et~al.(2021)He, Gao, and Chen}]{deberta}
Pengcheng He, Jianfeng Gao, and Weizhu Chen. 2021.
\newblock \href {http://arxiv.org/abs/2111.09543} {Debertav3: Improving deberta using electra-style pre-training with gradient-disentangled embedding sharing}.
\newblock \emph{CoRR}, abs/2111.09543.

\bibitem[{Hendrickx et~al.(2010)Hendrickx, Kim, Kozareva, Nakov, S{\'e}aghdha, Pad{\'o}, Pennacchiotti, Romano, and Szpakowicz}]{SemEval2010Task8}
Iris Hendrickx, Su~Nam Kim, Zornitsa Kozareva, Preslav Nakov, Diarmuid~{\'O} S{\'e}aghdha, Sebastian Pad{\'o}, Marco Pennacchiotti, Lorenza Romano, and Stan Szpakowicz. 2010.
\newblock Semeval-2010 task 8: Multi-way classification of semantic relations between pairs of nominals.
\newblock In \emph{*SEMEVAL}.

\bibitem[{Hendrycks and Gimpel(2023)}]{gelu}
Dan Hendrycks and Kevin Gimpel. 2023.
\newblock \href {https://doi.org/10.48550/arXiv.1606.08415} {Gaussian {Error} {Linear} {Units} ({GELUs})}.
\newblock ArXiv:1606.08415 [cs].

\bibitem[{Hovy et~al.(2006)Hovy, Marcus, Palmer, Ramshaw, and Weischedel}]{ontoNotes5}
Eduard~H. Hovy, Mitchell~P. Marcus, Martha Palmer, Lance~A. Ramshaw, and Ralph~M. Weischedel. 2006.
\newblock Ontonotes: The 90\% solution.
\newblock In \emph{North American Chapter of the Association for Computational Linguistics}.

\bibitem[{Huang et~al.(2019)Huang, Bras, Bhagavatula, and Choi}]{CosmosQA}
Lifu Huang, Ronan~Le Bras, Chandra Bhagavatula, and Yejin Choi. 2019.
\newblock Cosmos {QA:} machine reading comprehension with contextual commonsense reasoning.
\newblock In \emph{{EMNLP/IJCNLP} {(1)}}, pages 2391--2401. Association for Computational Linguistics.

\bibitem[{Jat et~al.(2018)Jat, Khandelwal, and Talukdar}]{GIDS}
Sharmistha Jat, Siddhesh Khandelwal, and Partha~Pratim Talukdar. 2018.
\newblock Improving distantly supervised relation extraction using word and entity based attention.
\newblock \emph{ArXiv}, abs/1804.06987.

\bibitem[{Jin et~al.(2020)Jin, Pan, Oufattole, Weng, Fang, and Szolovits}]{MedQA}
Di~Jin, Eileen Pan, Nassim Oufattole, Wei{-}Hung Weng, Hanyi Fang, and Peter Szolovits. 2020.
\newblock \href {http://arxiv.org/abs/2009.13081} {What disease does this patient have? {A} large-scale open domain question answering dataset from medical exams}.
\newblock \emph{CoRR}, abs/2009.13081.

\bibitem[{Jing et~al.(2019)Jing, Xiong, and Yan}]{BiPaR}
Yimin Jing, Deyi Xiong, and Zhen Yan. 2019.
\newblock \href {https://doi.org/10.18653/v1/D19-1249} {{B}i{P}a{R}: A bilingual parallel dataset for multilingual and cross-lingual reading comprehension on novels}.
\newblock In \emph{Proceedings of the 2019 Conference on Empirical Methods in Natural Language Processing and the 9th International Joint Conference on Natural Language Processing (EMNLP-IJCNLP)}, pages 2452--2462, Hong Kong, China. Association for Computational Linguistics.

\bibitem[{Josifoski et~al.(2022)Josifoski, De~Cao, Peyrard, Petroni, and West}]{genie}
Martin Josifoski, Nicola De~Cao, Maxime Peyrard, Fabio Petroni, and Robert West. 2022.
\newblock \href {https://doi.org/10.18653/v1/2022.naacl-main.342} {{G}en{IE}: Generative information extraction}.
\newblock In \emph{Proceedings of the 2022 Conference of the North American Chapter of the Association for Computational Linguistics: Human Language Technologies}, pages 4626--4643, Seattle, United States. Association for Computational Linguistics.

\bibitem[{Karimi et~al.(2015)Karimi, Metke{-}Jimenez, Kemp, and Wang}]{cadec}
Sarvnaz Karimi, Alejandro Metke{-}Jimenez, Madonna Kemp, and Chen Wang. 2015.
\newblock \href {https://doi.org/10.1016/j.jbi.2015.03.010} {Cadec: {A} corpus of adverse drug event annotations}.
\newblock \emph{J. Biomed. Informatics}, 55:73--81.

\bibitem[{Khashabi et~al.(2018)Khashabi, Chaturvedi, Roth, Upadhyay, and Roth}]{MultiRC}
Daniel Khashabi, Snigdha Chaturvedi, Michael Roth, Shyam Upadhyay, and Dan Roth. 2018.
\newblock \href {https://doi.org/10.18653/v1/N18-1023} {Looking beyond the surface: A challenge set for reading comprehension over multiple sentences}.
\newblock In \emph{Proceedings of the 2018 Conference of the North {A}merican Chapter of the Association for Computational Linguistics: Human Language Technologies, Volume 1 (Long Papers)}, pages 252--262, New Orleans, Louisiana. Association for Computational Linguistics.

\bibitem[{Khot et~al.(2020)Khot, Clark, Guerquin, Jansen, and Sabharwal}]{QASC}
Tushar Khot, Peter Clark, Michal Guerquin, Peter Jansen, and Ashish Sabharwal. 2020.
\newblock {QASC:} {A} dataset for question answering via sentence composition.
\newblock In \emph{{AAAI}}, pages 8082--8090. {AAAI} Press.

\bibitem[{Kim et~al.(2003)Kim, Ohta, Tateisi, and Tsujii}]{GENIA}
Jin-Dong Kim, Tomoko Ohta, Yuka Tateisi, and Jun'ichi Tsujii. 2003.
\newblock \href {https://doi.org/10.1093/bioinformatics/btg1023} {Genia corpus—a semantically annotated corpus for bio-textmining}.
\newblock \emph{Bioinformatics (Oxford, England)}, 19 Suppl 1:i180--2.

\bibitem[{Kocaman and Talby(2020)}]{bc2gm}
Veysel Kocaman and David Talby. 2020.
\newblock Biomedical named entity recognition at scale.
\newblock In \emph{ICPR Workshops}.

\bibitem[{Krallinger et~al.(2015)Krallinger, Rabal, Leitner, Vazquez, Salgado, lu, Leaman, Lu, Ji, Lowe, Sayle, Batista-Navarro, Rak, Huber, Rocktäschel, Matos, Campos, Tang, Qi, and Valencia}]{bc4chemd}
Martin Krallinger, Obdulia Rabal, Florian Leitner, Miguel Vazquez, David Salgado, Zhiyong lu, Robert Leaman, Yanan Lu, Donghong Ji, Daniel Lowe, Roger Sayle, Riza Batista-Navarro, Rafal Rak, Torsten Huber, Tim Rocktäschel, Sérgio Matos, David Campos, Buzhou Tang, Wang Qi, and Alfonso Valencia. 2015.
\newblock \href {https://doi.org/10.1186/1758-2946-7-S1-S2} {The chemdner corpus of chemicals and drugs and its annotation principles}.
\newblock \emph{Journal of Cheminformatics}, 7:S2.

\bibitem[{Kumar and Starly(2021)}]{FabNER}
Aman Kumar and Binil Starly. 2021.
\newblock “fabner”: information extraction from manufacturing process science domain literature using named entity recognition.
\newblock \emph{Journal of Intelligent Manufacturing}, 33:2393 -- 2407.

\bibitem[{Lai et~al.(2017)Lai, Xie, Liu, Yang, and Hovy}]{RACE}
Guokun Lai, Qizhe Xie, Hanxiao Liu, Yiming Yang, and Eduard~H. Hovy. 2017.
\newblock {RACE:} large-scale reading comprehension dataset from examinations.
\newblock In \emph{{EMNLP}}, pages 785--794. Association for Computational Linguistics.

\bibitem[{Lehmann et~al.(2015)Lehmann, Isele, Jakob, Jentzsch, Kontokostas, Mendes, Hellmann, Morsey, van Kleef, Auer, and Bizer}]{dbpedia}
Jens Lehmann, Robert Isele, Max Jakob, Anja Jentzsch, Dimitris Kontokostas, Pablo~N. Mendes, Sebastian Hellmann, Mohamed Morsey, Patrick van Kleef, S{\"{o}}ren Auer, and Christian Bizer. 2015.
\newblock \href {https://doi.org/10.3233/SW-140134} {Dbpedia - {A} large-scale, multilingual knowledge base extracted from wikipedia}.
\newblock \emph{Semantic Web}, 6(2):167--195.

\bibitem[{Levesque et~al.(2012)Levesque, Davis, and Morgenstern}]{WNLI}
Hector~J. Levesque, Ernest Davis, and Leora Morgenstern. 2012.
\newblock The winograd schema challenge.
\newblock In \emph{Proceedings of the Thirteenth International Conference on Principles of Knowledge Representation and Reasoning}, KR'12, page 552–561. AAAI Press.

\bibitem[{Lewis et~al.(2020)Lewis, Liu, Goyal, Ghazvininejad, Mohamed, Levy, Stoyanov, and Zettlemoyer}]{bart}
Mike Lewis, Yinhan Liu, Naman Goyal, Marjan Ghazvininejad, Abdelrahman Mohamed, Omer Levy, Veselin Stoyanov, and Luke Zettlemoyer. 2020.
\newblock \href {https://doi.org/10.18653/v1/2020.acl-main.703} {{BART}: Denoising sequence-to-sequence pre-training for natural language generation, translation, and comprehension}.
\newblock In \emph{Proceedings of the 58th Annual Meeting of the Association for Computational Linguistics}, pages 7871--7880, Online. Association for Computational Linguistics.

\bibitem[{Li et~al.(2016)Li, Sun, Johnson, Sciaky, Wei, Leaman, Davis, Mattingly, Wiegers, and Lu}]{bc5cdr}
Jiao Li, Yueping Sun, Robin~J. Johnson, Daniela Sciaky, Chih-Hsuan Wei, Robert Leaman, Allan~Peter Davis, Carolyn~J. Mattingly, Thomas~C. Wiegers, and Zhiyong Lu. 2016.
\newblock Biocreative v cdr task corpus: a resource for chemical disease relation extraction.
\newblock \emph{Database: The Journal of Biological Databases and Curation}, 2016.

\bibitem[{Li et~al.(2022)Li, Fei, Liu, Wu, Zhang, Teng, Ji, and Li}]{w2ner}
Jingye Li, Hao Fei, Jiang Liu, Shengqiong Wu, Meishan Zhang, Chong Teng, Donghong Ji, and Fei Li. 2022.
\newblock \href {https://ojs.aaai.org/index.php/AAAI/article/view/21344} {Unified named entity recognition as word-word relation classification}.
\newblock In \emph{Thirty-Sixth {AAAI} Conference on Artificial Intelligence, {AAAI} 2022, Thirty-Fourth Conference on Innovative Applications of Artificial Intelligence, {IAAI} 2022, The Twelveth Symposium on Educational Advances in Artificial Intelligence, {EAAI} 2022 Virtual Event, February 22 - March 1, 2022}, pages 10965--10973. {AAAI} Press.

\bibitem[{Lin et~al.(2020)Lin, Ji, Huang, and Wu}]{oneie}
Ying Lin, Heng Ji, Fei Huang, and Lingfei Wu. 2020.
\newblock \href {https://doi.org/10.18653/v1/2020.acl-main.713} {A joint neural model for information extraction with global features}.
\newblock In \emph{Proceedings of the 58th Annual Meeting of the Association for Computational Linguistics}, pages 7999--8009, Online. Association for Computational Linguistics.

\bibitem[{Liu et~al.(2013)Liu, Pasupat, Cyphers, and Glass}]{mit_ner_corpus}
Jingjing Liu, Panupong Pasupat, Scott Cyphers, and James~R. Glass. 2013.
\newblock \href {https://doi.org/10.1109/ICASSP.2013.6639301} {Asgard: {A} portable architecture for multilingual dialogue systems}.
\newblock In \emph{{IEEE} International Conference on Acoustics, Speech and Signal Processing, {ICASSP} 2013, Vancouver, BC, Canada, May 26-31, 2013}, pages 8386--8390. {IEEE}.

\bibitem[{Liu et~al.(2019)Liu, Ott, Goyal, Du, Joshi, Chen, Levy, Lewis, Zettlemoyer, and Stoyanov}]{roberta}
Yinhan Liu, Myle Ott, Naman Goyal, Jingfei Du, Mandar Joshi, Danqi Chen, Omer Levy, Mike Lewis, Luke Zettlemoyer, and Veselin Stoyanov. 2019.
\newblock \href {http://arxiv.org/abs/1907.11692} {Roberta: {A} robustly optimized {BERT} pretraining approach}.
\newblock \emph{CoRR}, abs/1907.11692.

\bibitem[{Liu et~al.(2021)Liu, Xu, Yu, Dai, Ji, Cahyawijaya, Madotto, and Fung}]{crossner}
Zihan Liu, Yan Xu, Tiezheng Yu, Wenliang Dai, Ziwei Ji, Samuel Cahyawijaya, Andrea Madotto, and Pascale Fung. 2021.
\newblock \href {https://ojs.aaai.org/index.php/AAAI/article/view/17587} {Crossner: Evaluating cross-domain named entity recognition}.
\newblock In \emph{Thirty-Fifth {AAAI} Conference on Artificial Intelligence, {AAAI} 2021, Thirty-Third Conference on Innovative Applications of Artificial Intelligence, {IAAI} 2021, The Eleventh Symposium on Educational Advances in Artificial Intelligence, {EAAI} 2021, Virtual Event, February 2-9, 2021}, pages 13452--13460. {AAAI} Press.

\bibitem[{Lou et~al.(2023)Lou, Lu, Dai, Jia, Lin, Han, Sun, and Wu}]{usm}
Jie Lou, Yaojie Lu, Dai Dai, Wei Jia, Hongyu Lin, Xianpei Han, Le~Sun, and Hua Wu. 2023.
\newblock \href {https://doi.org/10.48550/arXiv.2301.03282} {Universal information extraction as unified semantic matching}.
\newblock \emph{CoRR}, abs/2301.03282.

\bibitem[{Lu et~al.(2022)Lu, Liu, Dai, Xiao, Lin, Han, Sun, and Wu}]{uie}
Yaojie Lu, Qing Liu, Dai Dai, Xinyan Xiao, Hongyu Lin, Xianpei Han, Le~Sun, and Hua Wu. 2022.
\newblock \href {https://doi.org/10.18653/v1/2022.acl-long.395} {Unified structure generation for universal information extraction}.
\newblock In \emph{Proceedings of the 60th Annual Meeting of the Association for Computational Linguistics (Volume 1: Long Papers)}, pages 5755--5772, Dublin, Ireland. Association for Computational Linguistics.

\bibitem[{Luan et~al.(2018)Luan, He, Ostendorf, and Hajishirzi}]{scierc}
Yi~Luan, Luheng He, Mari Ostendorf, and Hannaneh Hajishirzi. 2018.
\newblock \href {https://doi.org/10.18653/v1/D18-1360} {Multi-task identification of entities, relations, and coreference for scientific knowledge graph construction}.
\newblock In \emph{Proceedings of the 2018 Conference on Empirical Methods in Natural Language Processing}, pages 3219--3232, Brussels, Belgium. Association for Computational Linguistics.

\bibitem[{Maas et~al.(2011)Maas, Daly, Pham, Huang, Ng, and Potts}]{IMDB}
Andrew~L. Maas, Raymond~E. Daly, Peter~T. Pham, Dan Huang, Andrew~Y. Ng, and Christopher Potts. 2011.
\newblock \href {http://www.aclweb.org/anthology/P11-1015} {Learning word vectors for sentiment analysis}.
\newblock In \emph{Proceedings of the 49th Annual Meeting of the Association for Computational Linguistics: Human Language Technologies}, pages 142--150, Portland, Oregon, USA. Association for Computational Linguistics.

\bibitem[{Mihaylov et~al.(2018)Mihaylov, Clark, Khot, and Sabharwal}]{OpenBookQA}
Todor Mihaylov, Peter Clark, Tushar Khot, and Ashish Sabharwal. 2018.
\newblock Can a suit of armor conduct electricity? {A} new dataset for open book question answering.
\newblock In \emph{{EMNLP}}, pages 2381--2391. Association for Computational Linguistics.

\bibitem[{Mitchell et~al.(2005)Mitchell, Strassel, Huang, and Zakhary}]{ace04}
Alexis Mitchell, Stephanie Strassel, Shudong Huang, and Ramez Zakhary. 2005.
\newblock \href {https://doi.org/https://doi.org/10.35111/8m4r-v312} {Ace 2004 multilingual training corpus}.

\bibitem[{Nguyen et~al.(2016)Nguyen, Rosenberg, Song, Gao, Tiwary, Majumder, and Deng}]{MS_MARCO}
Tri Nguyen, Mir Rosenberg, Xia Song, Jianfeng Gao, Saurabh Tiwary, Rangan Majumder, and Li~Deng. 2016.
\newblock \href {http://arxiv.org/abs/1611.09268} {{MS} {MARCO:} {A} human generated machine reading comprehension dataset}.
\newblock \emph{CoRR}, abs/1611.09268.

\bibitem[{Nie et~al.(2020)Nie, Williams, Dinan, Bansal, Weston, and Kiela}]{ANLI}
Yixin Nie, Adina Williams, Emily Dinan, Mohit Bansal, Jason Weston, and Douwe Kiela. 2020.
\newblock Adversarial nli: A new benchmark for natural language understanding.
\newblock In \emph{Proceedings of the 58th Annual Meeting of the Association for Computational Linguistics}. Association for Computational Linguistics.

\bibitem[{Pan et~al.(2017)Pan, Zhang, May, Nothman, Knight, and Ji}]{WikiAnn_en}
Xiaoman Pan, Boliang Zhang, Jonathan May, Joel Nothman, Kevin Knight, and Heng Ji. 2017.
\newblock \href {https://doi.org/10.18653/v1/P17-1178} {Cross-lingual name tagging and linking for 282 languages}.
\newblock In \emph{Proceedings of the 55th Annual Meeting of the Association for Computational Linguistics (Volume 1: Long Papers)}, pages 1946--1958, Vancouver, Canada. Association for Computational Linguistics.

\bibitem[{Paolini et~al.(2021)Paolini, Athiwaratkun, Krone, Ma, Achille, Anubhai, dos Santos, Xiang, and Soatto}]{tanl}
Giovanni Paolini, Ben Athiwaratkun, Jason Krone, Jie Ma, Alessandro Achille, Rishita Anubhai, C{\'{\i}}cero~Nogueira dos Santos, Bing Xiang, and Stefano Soatto. 2021.
\newblock \href {https://openreview.net/forum?id=US-TP-xnXI} {Structured prediction as translation between augmented natural languages}.
\newblock In \emph{9th International Conference on Learning Representations, {ICLR} 2021, Virtual Event, Austria, May 3-7, 2021}. OpenReview.net.

\bibitem[{Pontiki et~al.(2016)Pontiki, Galanis, Papageorgiou, Androutsopoulos, Manandhar, AL-Smadi, Al-Ayyoub, Zhao, Qin, De~Clercq, Hoste, Apidianaki, Tannier, Loukachevitch, Kotelnikov, Bel, Jim{\'e}nez-Zafra, and Eryi{\u{g}}it}]{absa16}
Maria Pontiki, Dimitris Galanis, Haris Papageorgiou, Ion Androutsopoulos, Suresh Manandhar, Mohammad AL-Smadi, Mahmoud Al-Ayyoub, Yanyan Zhao, Bing Qin, Orph{\'e}e De~Clercq, V{\'e}ronique Hoste, Marianna Apidianaki, Xavier Tannier, Natalia Loukachevitch, Evgeniy Kotelnikov, Nuria Bel, Salud~Mar{\'\i}a Jim{\'e}nez-Zafra, and G{\"u}l{\c{s}}en Eryi{\u{g}}it. 2016.
\newblock \href {https://doi.org/10.18653/v1/S16-1002} {{S}em{E}val-2016 task 5: Aspect based sentiment analysis}.
\newblock In \emph{Proceedings of the 10th International Workshop on Semantic Evaluation ({S}em{E}val-2016)}, pages 19--30, San Diego, California. Association for Computational Linguistics.

\bibitem[{Pontiki et~al.(2015)Pontiki, Galanis, Papageorgiou, Manandhar, and Androutsopoulos}]{absa15}
Maria Pontiki, Dimitris Galanis, Haris Papageorgiou, Suresh Manandhar, and Ion Androutsopoulos. 2015.
\newblock \href {https://doi.org/10.18653/v1/S15-2082} {{S}em{E}val-2015 task 12: Aspect based sentiment analysis}.
\newblock In \emph{Proceedings of the 9th International Workshop on Semantic Evaluation ({S}em{E}val 2015)}, pages 486--495, Denver, Colorado. Association for Computational Linguistics.

\bibitem[{Pontiki et~al.(2014)Pontiki, Galanis, Pavlopoulos, Papageorgiou, Androutsopoulos, and Manandhar}]{absa14}
Maria Pontiki, Dimitris Galanis, John Pavlopoulos, Harris Papageorgiou, Ion Androutsopoulos, and Suresh Manandhar. 2014.
\newblock \href {https://doi.org/10.3115/v1/S14-2004} {{S}em{E}val-2014 task 4: Aspect based sentiment analysis}.
\newblock In \emph{Proceedings of the 8th International Workshop on Semantic Evaluation ({S}em{E}val 2014)}, pages 27--35, Dublin, Ireland. Association for Computational Linguistics.

\bibitem[{Pyysalo and Ananiadou(2013)}]{AnatEM}
Sampo Pyysalo and Sophia Ananiadou. 2013.
\newblock \href {https://doi.org/10.1093/bioinformatics/btt580} {{Anatomical entity mention recognition at literature scale}}.
\newblock \emph{Bioinformatics}, 30(6):868--875.

\bibitem[{Qu et~al.(2023{\natexlab{a}})Qu, Gu, Xia, Li, Wang, and Huai}]{qu2023survey}
Xiaoye Qu, Yingjie Gu, Qingrong Xia, Zechang Li, Zhefeng Wang, and Baoxing Huai. 2023{\natexlab{a}}.
\newblock A survey on arabic named entity recognition: Past, recent advances, and future trends.
\newblock \emph{arXiv preprint arXiv:2302.03512}.

\bibitem[{Qu et~al.(2023{\natexlab{b}})Qu, Zeng, Liu, Wang, Huai, and Zhou}]{qu2023distantly}
Xiaoye Qu, Jun Zeng, Daizong Liu, Zhefeng Wang, Baoxing Huai, and Pan Zhou. 2023{\natexlab{b}}.
\newblock Distantly-supervised named entity recognition with adaptive teacher learning and fine-grained student ensemble.
\newblock In \emph{Proceedings of the AAAI Conference on Artificial Intelligence}, volume~37, pages 13501--13509.

\bibitem[{Raffel et~al.(2020)Raffel, Shazeer, Roberts, Lee, Narang, Matena, Zhou, Li, and Liu}]{t5}
Colin Raffel, Noam Shazeer, Adam Roberts, Katherine Lee, Sharan Narang, Michael Matena, Yanqi Zhou, Wei Li, and Peter~J. Liu. 2020.
\newblock \href {http://jmlr.org/papers/v21/20-074.html} {Exploring the limits of transfer learning with a unified text-to-text transformer}.
\newblock \emph{J. Mach. Learn. Res.}, 21:140:1--140:67.

\bibitem[{Rajani et~al.(2019)Rajani, McCann, Xiong, and Socher}]{cos_e}
Nazneen~Fatema Rajani, Bryan McCann, Caiming Xiong, and Richard Socher. 2019.
\newblock Explain yourself! leveraging language models for commonsense reasoning.
\newblock In \emph{{ACL} {(1)}}, pages 4932--4942. Association for Computational Linguistics.

\bibitem[{Rajpurkar et~al.(2018)Rajpurkar, Jia, and Liang}]{squad-v2}
Pranav Rajpurkar, Robin Jia, and Percy Liang. 2018.
\newblock \href {https://doi.org/10.18653/v1/P18-2124} {Know what you don{'}t know: Unanswerable questions for {SQ}u{AD}}.
\newblock In \emph{Proceedings of the 56th Annual Meeting of the Association for Computational Linguistics (Volume 2: Short Papers)}, pages 784--789, Melbourne, Australia. Association for Computational Linguistics.

\bibitem[{Riedel et~al.(2010)Riedel, Yao, and McCallum}]{nyt}
Sebastian Riedel, Limin Yao, and Andrew McCallum. 2010.
\newblock Modeling relations and their mentions without labeled text.
\newblock In \emph{Machine Learning and Knowledge Discovery in Databases}, pages 148--163, Berlin, Heidelberg. Springer Berlin Heidelberg.

\bibitem[{Roth and Yih(2004)}]{conll04}
Dan Roth and Wen-tau Yih. 2004.
\newblock \href {https://aclanthology.org/W04-2401} {A linear programming formulation for global inference in natural language tasks}.
\newblock In \emph{Proceedings of the Eighth Conference on Computational Natural Language Learning ({C}o{NLL}-2004) at {HLT}-{NAACL} 2004}, pages 1--8, Boston, Massachusetts, USA. Association for Computational Linguistics.

\bibitem[{Sakaguchi et~al.(2020)Sakaguchi, Bras, Bhagavatula, and Choi}]{Winogrande}
Keisuke Sakaguchi, Ronan~Le Bras, Chandra Bhagavatula, and Yejin Choi. 2020.
\newblock \href {https://ojs.aaai.org/index.php/AAAI/article/view/6399} {Winogrande: An adversarial winograd schema challenge at scale}.
\newblock In \emph{The Thirty-Fourth {AAAI} Conference on Artificial Intelligence, {AAAI} 2020, The Thirty-Second Innovative Applications of Artificial Intelligence Conference, {IAAI} 2020, The Tenth {AAAI} Symposium on Educational Advances in Artificial Intelligence, {EAAI} 2020, New York, NY, USA, February 7-12, 2020}, pages 8732--8740. {AAAI} Press.

\bibitem[{Satyapanich et~al.(2020)Satyapanich, Ferraro, and Finin}]{casie}
Taneeya Satyapanich, Francis Ferraro, and Tim Finin. 2020.
\newblock \href {https://doi.org/10.1609/aaai.v34i05.6401} {Casie: Extracting cybersecurity event information from text}.
\newblock In \emph{Proceedings of the AAAI Conference on Artificial Intelligence}, volume~34, pages 8749--8757.

\bibitem[{Socher et~al.(2013)Socher, Perelygin, Wu, Chuang, Manning, Ng, and Potts}]{sst-2}
Richard Socher, Alex Perelygin, Jean Wu, Jason Chuang, Christopher~D. Manning, Andrew~Y. Ng, and Christopher Potts. 2013.
\newblock \href {https://aclanthology.org/D13-1170/} {Recursive deep models for semantic compositionality over a sentiment treebank}.
\newblock In \emph{Proceedings of the 2013 Conference on Empirical Methods in Natural Language Processing, {EMNLP} 2013, 18-21 October 2013, Grand Hyatt Seattle, Seattle, Washington, USA, {A} meeting of SIGDAT, a Special Interest Group of the {ACL}}, pages 1631--1642. {ACL}.

\bibitem[{Srivastava et~al.(2014)Srivastava, Hinton, Krizhevsky, Sutskever, and Salakhutdinov}]{dropout}
Nitish Srivastava, Geoffrey~E. Hinton, Alex Krizhevsky, Ilya Sutskever, and Ruslan Salakhutdinov. 2014.
\newblock \href {https://doi.org/10.5555/2627435.2670313} {Dropout: a simple way to prevent neural networks from overfitting}.
\newblock \emph{J. Mach. Learn. Res.}, 15(1):1929--1958.

\bibitem[{Strauss et~al.(2016)Strauss, Toma, Ritter, de~Marneffe, and Xu}]{WNUT-16}
Benjamin Strauss, Bethany Toma, Alan Ritter, Marie-Catherine de~Marneffe, and Wei Xu. 2016.
\newblock \href {https://aclanthology.org/W16-3919} {Results of the {WNUT}16 named entity recognition shared task}.
\newblock In \emph{Proceedings of the 2nd Workshop on Noisy User-generated Text ({WNUT})}, pages 138--144, Osaka, Japan. The COLING 2016 Organizing Committee.

\bibitem[{Su et~al.(2021)Su, Lu, Pan, Wen, and Liu}]{roformer}
Jianlin Su, Yu~Lu, Shengfeng Pan, Bo~Wen, and Yunfeng Liu. 2021.
\newblock \href {http://arxiv.org/abs/2104.09864} {Roformer: Enhanced transformer with rotary position embedding}.
\newblock \emph{CoRR}, abs/2104.09864.

\bibitem[{Su et~al.(2022)Su, Murtadha, Pan, Hou, Sun, Huang, Wen, and Liu}]{global_pointer}
Jianlin Su, Ahmed Murtadha, Shengfeng Pan, Jing Hou, Jun Sun, Wanwei Huang, Bo~Wen, and Yunfeng Liu. 2022.
\newblock \href {https://doi.org/10.48550/arXiv.2208.03054} {Global {Pointer}: {Novel} {Efficient} {Span}-based {Approach} for {Named} {Entity} {Recognition}}.
\newblock ArXiv:2208.03054 [cs].

\bibitem[{Sun et~al.(2019)Sun, Yu, Chen, Yu, Choi, and Cardie}]{DREAM}
Kai Sun, Dian Yu, Jianshu Chen, Dong Yu, Yejin Choi, and Claire Cardie. 2019.
\newblock {DREAM:} {A} challenge dataset and models for dialogue-based reading comprehension.
\newblock \emph{Trans. Assoc. Comput. Linguistics}, 7:217--231.

\bibitem[{Sun et~al.(2022)Sun, Li, Pergola, Wallace, John, Greene, Kim, and He}]{PHEE}
Zhao-Li Sun, Jiazheng Li, Gabriele Pergola, Byron~C. Wallace, Bino John, Nigel Greene, Joseph Kim, and Yulan He. 2022.
\newblock Phee: A dataset for pharmacovigilance event extraction from text.
\newblock \emph{ArXiv}, abs/2210.12560.

\bibitem[{Takanobu et~al.(2018)Takanobu, Zhang, Liu, and Huang}]{NYT11HRL}
Ryuichi Takanobu, Tianyang Zhang, Jiexi Liu, and Minlie Huang. 2018.
\newblock A hierarchical framework for relation extraction with reinforcement learning.
\newblock In \emph{AAAI Conference on Artificial Intelligence}.

\bibitem[{Tedeschi and Navigli(2022)}]{MultiNERD}
Simone Tedeschi and Roberto Navigli. 2022.
\newblock \href {https://doi.org/10.18653/v1/2022.findings-naacl.60} {{M}ulti{NERD}: A multilingual, multi-genre and fine-grained dataset for named entity recognition (and disambiguation)}.
\newblock In \emph{Findings of the Association for Computational Linguistics: NAACL 2022}, pages 801--812, Seattle, United States. Association for Computational Linguistics.

\bibitem[{Therasa and Mathivanan(2022)}]{mrc_survey}
M.~Therasa and G.~Mathivanan. 2022.
\newblock \href {https://doi.org/10.1109/ICCMC53470.2022.9754070} {Survey of machine reading comprehension models and its evaluation metrics}.
\newblock In \emph{2022 6th International Conference on Computing Methodologies and Communication (ICCMC)}, pages 1006--1013.

\bibitem[{Tjong Kim~Sang and De~Meulder(2003)}]{conll03}
Erik~F. Tjong Kim~Sang and Fien De~Meulder. 2003.
\newblock \href {https://aclanthology.org/W03-0419/} {Introduction to the conll-2003 shared task: Language-independent named entity recognition}.
\newblock In \emph{Proceedings of CoNLL-2003}, pages 142--147. Edmonton, Canada.

\bibitem[{Trischler et~al.(2016)Trischler, Wang, Yuan, Harris, Sordoni, Bachman, and Suleman}]{NewsQA}
Adam Trischler, Tong Wang, Xingdi Yuan, Justin Harris, Alessandro Sordoni, Philip Bachman, and Kaheer Suleman. 2016.
\newblock \href {http://arxiv.org/abs/1611.09830} {Newsqa: {A} machine comprehension dataset}.
\newblock \emph{CoRR}, abs/1611.09830.

\bibitem[{Ushio et~al.(2022)Ushio, Neves, Silva, Barbieri, and Camacho-Collados}]{TweetNER7}
Asahi Ushio, Leonardo Neves, Vitor Silva, Francesco Barbieri, and Jose Camacho-Collados. 2022.
\newblock \href {http://arxiv.org/abs/2210.03797} {Named entity recognition in twitter: A dataset and analysis on short-term temporal shifts}.

\bibitem[{Wadden et~al.(2019)Wadden, Wennberg, Luan, and Hajishirzi}]{dygiepp}
David Wadden, Ulme Wennberg, Yi~Luan, and Hannaneh Hajishirzi. 2019.
\newblock \href {https://doi.org/10.18653/v1/D19-1585} {Entity, relation, and event extraction with contextualized span representations}.
\newblock In \emph{Proceedings of the 2019 Conference on Empirical Methods in Natural Language Processing and the 9th International Joint Conference on Natural Language Processing (EMNLP-IJCNLP)}, pages 5784--5789, Hong Kong, China. Association for Computational Linguistics.

\bibitem[{Walker et~al.(2006)Walker, Strassel, Medero, and Maeda}]{ace05}
Christopher Walker, Stephanie Strassel, Julie Medero, and Kazuaki Maeda. 2006.
\newblock \href {https://doi.org/https://doi.org/10.35111/mwxc-vh88} {Ace 2005 multilingual training corpus}.

\bibitem[{Wang et~al.(2019)Wang, Singh, Michael, Hill, Levy, and Bowman}]{glue}
Alex Wang, Amanpreet Singh, Julian Michael, Felix Hill, Omer Levy, and Samuel~R. Bowman. 2019.
\newblock \href {https://openreview.net/forum?id=rJ4km2R5t7} {{GLUE:} {A} multi-task benchmark and analysis platform for natural language understanding}.
\newblock In \emph{7th International Conference on Learning Representations, {ICLR} 2019, New Orleans, LA, USA, May 6-9, 2019}. OpenReview.net.

\bibitem[{Wang et~al.(2022)Wang, Liu, Chen, Hong, Tang, and Song}]{deepstruct}
Chenguang Wang, Xiao Liu, Zui Chen, Haoyun Hong, Jie Tang, and Dawn Song. 2022.
\newblock \href {https://doi.org/10.18653/v1/2022.findings-acl.67} {{D}eep{S}truct: Pretraining of language models for structure prediction}.
\newblock In \emph{Findings of the Association for Computational Linguistics: ACL 2022}, pages 803--823, Dublin, Ireland. Association for Computational Linguistics.

\bibitem[{Wang et~al.(2023)Wang, Zhou, Zu, Xia, Chen, Zhang, Zheng, Ye, Zhang, Gui, Kang, Yang, Li, and Du}]{instructuie}
Xiao Wang, Weikang Zhou, Can Zu, Han Xia, Tianze Chen, Yuansen Zhang, Rui Zheng, Junjie Ye, Qi~Zhang, Tao Gui, Jihua Kang, Jingsheng Yang, Siyuan Li, and Chunsai Du. 2023.
\newblock \href {https://doi.org/10.48550/arXiv.2304.08085} {{InstructUIE}: {Multi}-task {Instruction} {Tuning} for {Unified} {Information} {Extraction}}.
\newblock ArXiv:2304.08085 [cs].

\bibitem[{Wang et~al.(2021)Wang, Yu, Zhu, Liu, Yu, and Sun}]{mac-discontinuous-ner}
Yucheng Wang, Bowen Yu, Hongsong Zhu, Tingwen Liu, Nan Yu, and Limin Sun. 2021.
\newblock \href {https://doi.org/10.18653/v1/2021.acl-long.63} {Discontinuous named entity recognition as maximal clique discovery}.
\newblock In \emph{Proceedings of the 59th Annual Meeting of the Association for Computational Linguistics and the 11th International Joint Conference on Natural Language Processing (Volume 1: Long Papers)}, pages 764--774, Online. Association for Computational Linguistics.

\bibitem[{Warstadt et~al.(2019)Warstadt, Singh, and Bowman}]{cola}
Alex Warstadt, Amanpreet Singh, and Samuel~R. Bowman. 2019.
\newblock \href {https://doi.org/10.1162/tacl\_a\_00290} {Neural network acceptability judgments}.
\newblock \emph{Trans. Assoc. Comput. Linguistics}, 7:625--641.

\bibitem[{Welbl et~al.(2017)Welbl, Liu, and Gardner}]{SciQ}
Johannes Welbl, Nelson~F. Liu, and Matt Gardner. 2017.
\newblock \href {https://doi.org/10.18653/v1/W17-4413} {Crowdsourcing multiple choice science questions}.
\newblock In \emph{Proceedings of the 3rd Workshop on Noisy User-generated Text}, pages 94--106, Copenhagen, Denmark. Association for Computational Linguistics.

\bibitem[{Williams et~al.(2018)Williams, Nangia, and Bowman}]{mnli}
Adina Williams, Nikita Nangia, and Samuel~R. Bowman. 2018.
\newblock \href {https://doi.org/10.18653/v1/n18-1101} {A broad-coverage challenge corpus for sentence understanding through inference}.
\newblock In \emph{Proceedings of the 2018 Conference of the North American Chapter of the Association for Computational Linguistics: Human Language Technologies, {NAACL-HLT} 2018, New Orleans, Louisiana, USA, June 1-6, 2018, Volume 1 (Long Papers)}, pages 1112--1122. Association for Computational Linguistics.

\bibitem[{Yan et~al.(2021{\natexlab{a}})Yan, Dai, Ji, Qiu, and Zhang}]{bart-absa}
Hang Yan, Junqi Dai, Tuo Ji, Xipeng Qiu, and Zheng Zhang. 2021{\natexlab{a}}.
\newblock \href {https://doi.org/10.18653/v1/2021.acl-long.188} {A unified generative framework for aspect-based sentiment analysis}.
\newblock In \emph{Proceedings of the 59th Annual Meeting of the Association for Computational Linguistics and the 11th International Joint Conference on Natural Language Processing (Volume 1: Long Papers)}, pages 2416--2429, Online. Association for Computational Linguistics.

\bibitem[{Yan et~al.(2021{\natexlab{b}})Yan, Gui, Dai, Guo, Zhang, and Qiu}]{bart-ner}
Hang Yan, Tao Gui, Junqi Dai, Qipeng Guo, Zheng Zhang, and Xipeng Qiu. 2021{\natexlab{b}}.
\newblock \href {https://doi.org/10.18653/v1/2021.acl-long.451} {A unified generative framework for various {NER} subtasks}.
\newblock In \emph{Proceedings of the 59th Annual Meeting of the Association for Computational Linguistics and the 11th International Joint Conference on Natural Language Processing (Volume 1: Long Papers)}, pages 5808--5822, Online. Association for Computational Linguistics.

\bibitem[{Yang et~al.(2022)Yang, Wang, Gan, Zhu, Zhang, Wu, Gao, Zhang, and Sakai}]{unimc}
Ping Yang, Junjie Wang, Ruyi Gan, Xinyu Zhu, Lin Zhang, Ziwei Wu, Xinyu Gao, Jiaxing Zhang, and Tetsuya Sakai. 2022.
\newblock \href {https://aclanthology.org/2022.emnlp-main.474} {Zero-shot learners for natural language understanding via a unified multiple choice perspective}.
\newblock In \emph{Proceedings of the 2022 Conference on Empirical Methods in Natural Language Processing}, pages 7042--7055, Abu Dhabi, United Arab Emirates. Association for Computational Linguistics.

\bibitem[{Yu et~al.(2020)Yu, Jiang, Dong, and Feng}]{ReClor}
Weihao Yu, Zihang Jiang, Yanfei Dong, and Jiashi Feng. 2020.
\newblock \href {http://arxiv.org/abs/2002.04326} {Reclor: {A} reading comprehension dataset requiring logical reasoning}.
\newblock \emph{CoRR}, abs/2002.04326.

\bibitem[{Zellers et~al.(2019)Zellers, Holtzman, Bisk, Farhadi, and Choi}]{hellaswag}
Rowan Zellers, Ari Holtzman, Yonatan Bisk, Ali Farhadi, and Yejin Choi. 2019.
\newblock \href {https://doi.org/10.18653/v1/p19-1472} {Hellaswag: Can a machine really finish your sentence?}
\newblock In \emph{Proceedings of the 57th Conference of the Association for Computational Linguistics, {ACL} 2019, Florence, Italy, July 28- August 2, 2019, Volume 1: Long Papers}, pages 4791--4800. Association for Computational Linguistics.

\bibitem[{Zhang and Wang(2015)}]{KBP37}
Dongxu Zhang and Dong Wang. 2015.
\newblock \href {http://arxiv.org/abs/1508.01006} {Relation classification via recurrent neural network}.
\newblock \emph{CoRR}, abs/1508.01006.

\bibitem[{Zhang et~al.(2015)Zhang, Zhao, and LeCun}]{ag_news}
Xiang Zhang, Junbo~Jake Zhao, and Yann LeCun. 2015.
\newblock Character-level convolutional networks for text classification.
\newblock In \emph{{NIPS}}, pages 649--657.

\bibitem[{Zhu et~al.(2022)Zhu, Qu, Chen, Wang, Huai, Yuan, and Zhang}]{ptpcg}
Tong Zhu, Xiaoye Qu, Wenliang Chen, Zhefeng Wang, Baoxing Huai, Nicholas Yuan, and Min Zhang. 2022.
\newblock \href {https://doi.org/10.24963/ijcai.2022/632} {Efficient document-level event extraction via pseudo-trigger-aware pruned complete graph}.
\newblock In \emph{Proceedings of the Thirty-First International Joint Conference on Artificial Intelligence, {IJCAI-22}}, pages 4552--4558. International Joint Conferences on Artificial Intelligence Organization.
\newblock Main Track.

\end{thebibliography}
\bibliographystyle{acl_natbib}

\appendix
\newpage

\section{Comparisons on Information Indexing Strategies}\label{sec:indexing}

UIE~\cite{uie} provides the extracted information's positions based on string matching.
However, this strategy is not accurate and contains ambiguities.
To investigate the matching accuracy, we take the NER task as an example and use golden entity strings to calculate the upper bound F1 scores of different UIE string matching strategies. The table below shows that the upper bounds are quite low on the datasets (<30\%). This indicates that obtaining positions via string matching is ineffective and has serious ambiguity problems.

\begin{table}[h]
    \centering
    \resizebox{\columnwidth}{!}{
    \begin{tabular}{lrrr}
    \toprule
        NER & ACE04 & ACE05 & CoNLL03 \\
    \midrule
        Mirror & 100.00 & 100.00 & 100.00 \\
        UIE-first & 13.31 & 14.51 & 27.67 \\
        UIE-longer-first & 14.55 & 16.21 & 27.97 \\
    \bottomrule
    \end{tabular}
    }
    \caption{Upper bound of different string matching strategies on NER.}
    \label{tab:indexing_ner}
\end{table}

TANL~\cite{tanl} can provides exact positions in NER since it generates the enclosure tags.
However, it still faces the ambiguity problem when two entities have the same string in joint entity relation extraction because the tail entity is a generated text corresponding to an enclosed head entity (refer to section 3 in the TANL paper).
We also calculate the upper bound F1 scores of relation extraction in a TANL manner, and the results show it does not ideally generate perfect positions.

\begin{table}[h]
    \centering
    \resizebox{\columnwidth}{!}{
    \begin{tabular}{lrrrr}
    \toprule
    RE & ACE05 & CoNLL04 & NYT & SciERC \\
    \midrule
    Mirror & 100.00 & 100.00 & 100.00 & 100.00 \\
    TANL & 96.91 & 96.95 & 100.00 & 99.64 \\
    \bottomrule
    \end{tabular}
    }
    \caption{Upper bound of relation extraction with Mirror and TANL position indexing strategies.}
    \label{tab:indexing_rel}
\end{table}

\section{Hyper-parameter Settings}
\label{sec:app-hyper-param}

Table~\ref{tab:hyper-param} shows the hyper-parameters in our experiments.
For few-shot experiments, we follow \citet{uie} and generate 1-, 5-, 10-shot data with 5 seeds.

\begin{table}[h]
    \centering
    \begin{tabular}{cc}
        \toprule
        Item & Setting \\
        \midrule
            warmup proportion & 0.1 \\
            pretraining epochs & 3 \\
            fine-tuning epochs & 20 \\
            fine-tuning epoch patience & 3 \\
            few-shot epochs & 200 \\
            few-shot epoch patience & 10 \\
            batch size & 8 \\
            PLM learning rate & 2e-5 \\
            PLM weight decay & 0.1 \\
            others learning rate & 1e-4 \\
            max gradient norm & 1.0 \\
            $d_h$ & 1024 \\
            $d_b$ & 512 \\
            dropout & 0.3 \\
        \bottomrule
    \end{tabular}
    \caption{
        Hyper-parameter settings.
    }
    \label{tab:hyper-param}
\end{table}

\section{Dataset Statistics}
\label{sec:appendix_data}

This section contains detailed statistics for pretraining datasets and fine-tuning datasets.
Pretraining data statistics are listed in Table~\ref{tab:data-statistics-cls}, \ref{tab:data-statistics-ner}, \ref{tab:data-statistics-re}, \ref{tab:data-statistics-mrc} and \ref{tab:data-statistics-ee}.
For the sampling number $N_{\max}$ of each kind of dataset, please refer to Table~\ref{tab:pretrain-dataset-statistics}.
When collecting pretraining data, we refer to the datasets mentioned in \citet{mrc_survey} and \citet{unimc}.
Downstream data statistics are listed in Table~\ref{tab:downstream-data-statistics}.
We also provide direct inference results with the pretrained Mirror model in Table~\ref{tab:downstream-data-statistics}.

\begin{table}[t]
    \centering
    \resizebox{\columnwidth}{!}{
    \begin{tabular}{lcrr}
        \toprule
        Name & Citation & \#Instruction & \#Instance \\
        \midrule
        AG News & \cite{ag_news} & 5 & 5,000 \\
        ANLI$^{\clubsuit}$ & \cite{ANLI} & 29 & 15,000 \\
        ARC & \cite{ARC} & 3,361 & 3,370 \\
        CoLA & \cite{cola} & 43 & 5,000 \\
        CosmosQA & \cite{CosmosQA} & 4,483 & 5,000 \\
        CoS-E & \cite{cos_e} & 5,000 & 5,000 \\
        DBpedia & \cite{dbpedia} & 6 & 5,000 \\
        DREAM & \cite{DREAM} & 3,842 & 5,000 \\
        hellaswag & \cite{hellaswag} & 20 & 5,000 \\
        IMDB & \cite{IMDB} & 26 & 5,000 \\
        MedQA & \cite{MedQA} & 5,000 & 5,000 \\
        MNLI & \cite{mnli} & 29 & 5,000 \\
        MRPC & \cite{mrpc} & 40 & 3,668 \\
        MultiRC & \cite{MultiRC} & 4,999 & 5,000 \\
        OpenBookQA & \cite{OpenBookQA} & 4,835 & 4,957 \\
        QASC & \cite{QASC} & 4,832 & 5,000 \\
        QNLI & \cite{glue} & 31 & 5,000 \\
        QQP & \cite{glue} & 40 & 5,000 \\
        RACE & \cite{RACE} & 4,482 & 5,000 \\
        RACE-C & \cite{RACE} & 4,782 & 5,000 \\
        ReClor & \cite{ReClor} & 3,368 & 4,638 \\
        RTE & \cite{glue} & 29 & 2,490 \\
        SciQ & \cite{SciQ} & 4,989 & 5,000 \\
        SNLI & \cite{SNLI} & 29 & 5,000 \\
        SST-2 & \cite{sst-2} & 26 & 5,000 \\
        Winogrande & \cite{Winogrande} & 20 & 5,000 \\
        WNLI & \cite{WNLI} & 31 & 635 \\
        \midrule
        Total & & 54,070 & 134,758 \\
        \bottomrule
    \end{tabular}
    }
    \caption{
        Pretraining data statistics on classification.
        The maximal sampling number $N_{\max}$ for each dataset is 5,000.
        $^{\clubsuit}$: ANLI contains 3 subsets, so the total number is greater than 5,000.
    }
    \label{tab:data-statistics-cls}
\end{table}

\begin{table}[t]
    \centering
    \resizebox{\columnwidth}{!}{
    \begin{tabular}{lcrr}
        \toprule
        Name & Citation & \#Instruction & \#Instance \\
        \midrule
        BiPaR & \cite{BiPaR} & 11,524 & 11,668 \\
        MS MARCO v2.1 & \cite{MS_MARCO} & 20,000 & 20,000 \\
        NewsQA & \cite{NewsQA} & 19,659 & 20,000 \\
        SQuAD v2 & \cite{squad-v2} & 19,998 & 20,000 \\
        SubjQA & \cite{SubjQA} & 4,060 & 13,990 \\
        \midrule
        Total & & 75,220 & 85,658 \\
        \bottomrule
    \end{tabular}
    }
    \caption{
        Pretraining data statistics on MRC.
        The maximal sampling number $N_{\max}$ for each dataset is 20,000.
    }
    \label{tab:data-statistics-mrc}
\end{table}

\begin{table}[t]
    \centering
    \resizebox{0.8\columnwidth}{!}{
    \begin{tabular}{lcrr}
        \toprule
        Name & Citation &  \#Instruction & \#Instance \\
        \midrule
        PHEE & \cite{PHEE} & 40 & 2,898 \\
        \midrule
        Total & & 40 & 2,898 \\
        \bottomrule
    \end{tabular}
    }
    \caption{
        Pretraining data statistics on EE.
        Due to the scarcity of EE datasets, we sample all the instances ($N_{\max}=\infty$).
    }
    \label{tab:data-statistics-ee}
\end{table}

\begin{table}[t]
    \centering
    \resizebox{\columnwidth}{!}{
    \begin{tabular}{lcrr}
        \toprule
        Name & Citation & \#Instruction & \#Instance \\
        \midrule
        AnatEM & \cite{AnatEM} & 42 & 5,861 \\
        bc2gm & \cite{bc2gm} & 42 & 12,500 \\
        bc4chemd & \cite{bc4chemd} & 42 & 20,000 \\
        bc5cdr & \cite{bc5cdr} & 42 & 4,560 \\
        Broad Tweet Corpus & \cite{Broad_Twitter_Corpus} & 42 & 5,334 \\
        FabNER & \cite{FabNER} & 42 & 9,435 \\
        FindVehicle & \cite{FindVehicle} & 42 & 20,000 \\
        GENIA & \cite{GENIA} & 42 & 15,023 \\
        HarveyNER & \cite{HarveyNER} & 42 & 3,967 \\
        MultiNERD & \cite{MultiNERD} & 42 & 20,000 \\
        NCBIDisease & \cite{NCBIDisease} & 42 & 5,432 \\
        OntoNotes5 & \cite{ontoNotes5} & 42 & 20,000 \\
        TweetNER7 & \cite{TweetNER7} & 42 & 7,103 \\
        WikiANN\_en & \cite{WikiAnn_en} & 42 & 20,000 \\
        WNUT-16 & \cite{WNUT-16} & 42 & 2,394 \\
        \midrule
        Total & & 42 & 171,609 \\
        \bottomrule
    \end{tabular}
    }
    \caption{
        Pretraining data statistics on NER.
        The maximal sampling number $N_{\max}$ for each dataset is 20,000.
    }
    \label{tab:data-statistics-ner}
\end{table}

\begin{table}[t]
    \centering
    \resizebox{\columnwidth}{!}{
    \begin{tabular}{lcrr}
        \toprule
        Name & Citation & \#Instruction & \#Instance \\
        \midrule
        ADE & \cite{ADE} &  9 & 3,417 \\
        FewRel & \cite{FewRel} &  9 & 20,000 \\
        GIDS & \cite{GIDS} &  9 & 8,526 \\
        kbp37 & \cite{KBP37} &  9 & 15,807 \\
        NYT10 & \cite{nyt} &  9 & 20,000 \\
        NYT11HRL & \cite{NYT11HRL} &  9 & 20,000 \\
        SemEval2010 Task8 & \cite{SemEval2010Task8} &  9 & 8,000 \\
        WebNLG & \cite{WebNLG} &  9 & 5,019 \\
        Wiki-ZSL & \cite{Wiki-ZSL} &  9 & 23,107 \\
        \midrule
        Total & & 9 & 123,876 \\
        \bottomrule
    \end{tabular}
    }
    \caption{
        Pretraining data statistics on RE.
        The maximal sampling number $N_{\max}$ for each dataset is 20,000.
    }
    \label{tab:data-statistics-re}
\end{table}

\begin{table*}[t]
\centering
\resizebox{\textwidth}{!}{%
\begin{tabular}{ccccrrrcc}
\toprule
Task                            & Dataset    & Citation               & Metric                            & \#Train & \#Dev  & \#Test & Included in PT               & Mirror$_{\text{direct}}$ \\
\midrule
\multirow{3}{*}{NER}            & ACE04      & \citet{ace04}          & Entity Micro F1                   & 6,202   & 745    & 812    & \textcolor{red}{\xmark}      & 21.49                    \\
                                & ACE05      & \citet{ace05}          & Entity Micro F1                   & 7,299   & 971    & 1,060  & \textcolor{red}{\xmark}      & 18.70                    \\
                                & CoNLL03    & \citet{conll03}        & Entity Micro F1                   & 14,041  & 3,250  & 3,453  & \textcolor{red}{\xmark}      & 66.91                    \\
                                \midrule
\multirow{4}{*}{RE}             & ACE05      & \citet{ace05}          & Triplet Micro F1                  & 10,051  & 2,420  & 2,050  & \textcolor{red}{\xmark}      & 0.51                     \\
                                & CoNLL04    & \citet{conll04}        & Triplet Micro F1                  & 922     & 231    & 288    & \textcolor{red}{\xmark}      & 1.40                     \\
                                & NYT        & \citet{nyt}            & Triplet Micro F1                  & 56,196  & 5,000  & 5,000  & \textcolor{red}{\xmark}      & 69.67                    \\
                                & SciERC     & \citet{scierc}         & Triplet Micro F1                  & 1,861   & 275    & 551    & \textcolor{red}{\xmark}      & 0.00                     \\
                                \midrule
\multirow{2}{*}{EE}             & ACE05      & \citet{ace05}          & Trigger \& Argument Micro F1      & 19,216  & 901    & 676    & \textcolor{red}{\xmark}      & 3.99/0.00                \\
                                & CASIE      & \citet{casie}          & Trigger \& Argument Micro F1      & 11,189  & 1,778  & 3,208  & \textcolor{red}{\xmark}      & 2.13/0.00                \\
                                \midrule
\multirow{4}{*}{ABSA}           & 14-res     & \citet{absa14}         & Triplet Micro F1                  & 1,266   & 310    & 492    & \textcolor{red}{\xmark}      & 0.00                     \\
                                & 14-lap     & \citet{absa14}         & Triplet Micro F1                  & 906     & 219    & 328    & \textcolor{red}{\xmark}      & 0.00                     \\
                                & 15-res     & \citet{absa15}         & Triplet Micro F1                  & 605     & 148    & 322    & \textcolor{red}{\xmark}      & 0.00                     \\
                                & 16-res     & \citet{absa16}         & Triplet Micro F1                  & 857     & 210    & 326    & \textcolor{red}{\xmark}      & 0.00                     \\
                                \midrule
Discontinuous NER               & CADEC      & \citet{cadec}          & Entity Micro F1                   & 5,340   & 1,097  & 1,160  & \textcolor{red}{\xmark}      & 52.34                    \\
\midrule
Hyper RE                        & HyperRED   & \citet{hyperred}       & Tuple Micro F1                    & 39,840  & 4,000  & 1,000  & \textcolor{red}{\xmark}      & 0.00                     \\
\midrule
\multirow{7}{*}{Zero-shot NER}  & Movie      & \citet{mit_ner_corpus} & Entity Micro F1                   & 9,774   & 2,442  & 2,442  & \textcolor{red}{\xmark}      & 39.24                    \\
                                & Restaurant & \citet{mit_ner_corpus} & Entity Micro F1                   & 7,659   & 1,520  & 1,520  & \textcolor{red}{\xmark}      & 16.17                    \\
                                & AI         & \citet{crossner}       & Entity Micro F1                   & 100     & 350    & 431    & \textcolor{red}{\xmark}      & 45.91                    \\
                                & Literature & \citet{crossner}       & Entity Micro F1                   & 100     & 400    & 416    & \textcolor{red}{\xmark}      & 46.77                    \\
                                & Music      & \citet{crossner}       & Entity Micro F1                   & 100     & 380    & 465    & \textcolor{red}{\xmark}      & 59.12                    \\
                                & Politics   & \citet{crossner}       & Entity Micro F1                   & 199     & 540    & 650    & \textcolor{red}{\xmark}      & 67.27                    \\
                                & Science    & \citet{crossner}       & Entity Micro F1                   & 200     & 450    & 543    & \textcolor{red}{\xmark}      & 54.42                    \\
                                \midrule
MRC                             & SQuAD v2.0 & \citet{squad-v2}       & Exact Match \& F1                 & 86,821  & 5,928  & -      & {\color[HTML]{008114}\cmark} & 40.35/67.39              \\
\midrule
\multirow{7}{*}{Classification} & CoLA       & \citet{cola}           & Matthew's Correlation Coefficient & 8,551   & 527    & -      & {\color[HTML]{008114}\cmark} & 63.91                    \\
                                & QQP        & \citet{glue}           & Accuracy                          & 363,846 & 40,430 & -      & {\color[HTML]{008114}\cmark} & 84.84                    \\
                                & MNLI       & \citet{mnli}           & Accuracy                          & 392,702 & 9,815  & -      & {\color[HTML]{008114}\cmark} & 85.90                    \\
                                & SST-2      & \citet{sst-2}          & Accuracy                          & 67,350  & 873    & -      & {\color[HTML]{008114}\cmark} & 93.58                    \\
                                & QNLI       & \citet{glue}           & Accuracy                          & 104,743 & 5,463  & -      & {\color[HTML]{008114}\cmark} & 91.62                    \\
                                & RTE        & \citet{glue}           & Accuracy                          & 2,490   & 277    & -      & {\color[HTML]{008114}\cmark} & 85.92                    \\
                                & MRPC       & \citet{mrpc}           & Accuracy                          & 3,668   & 408    & 1,725  & {\color[HTML]{008114}\cmark} & 89.22 \\
                                \bottomrule
\end{tabular}%
}
\caption{Data statistics on downstream tasks. Included in PT stands for whether the dataset is included in the data pretraining corpus. Mirror$_{\text{direct}}$ is the model trained on the pretraining corpus.}
\label{tab:downstream-data-statistics}
\end{table*}

\section{Case Study}

We provide some interesting cases across different tasks with the pretrained Mirror w/ Inst. to manually evaluate its versatility on various tasks under zero-shot settings.
The model inputs \& outputs are presented in Table~\ref{tab:case-study}.

\begin{table*}[t]
    \centering
    \begin{tabular}{lp{12cm}}
        \toprule
        \multicolumn{2}{l}{\textit{Classification (Multi-choice MRC)}} \\
        \textbf{Input} & \verb|[I]| Mirror Mirror on the wall, who's the fairest of them all? \verb|[LC]| Evil Queen \verb|[LC]| Snow White \\
        \textbf{Output} & \verb|[LC]|$_{\text{Snow White}}$ \\
        \midrule
        \multicolumn{2}{l}{\textit{Extractive MRC}} \\
        \textbf{Input} & \verb|[I]| Mirror Mirror on the wall, who's the fairest of them all? \verb|[TP]| Evil Queen is jealous of Snow White's beauty. \\
        \textbf{Output} & \texttt{Snow White} \\
        \midrule
        \multicolumn{2}{l}{\textit{Named Entity Extraction}} \\
        \textbf{Input} & \verb|[I]| Mirror Mirror, please help me extract all the model names. \verb|[LM]| model name \verb|[TL]| LLaMA and OPT are open-sourced large language models. \\
        \textbf{Output} & \verb|[LM]|$_{\text{LLaMA}}$, \verb|[LM]|$_{\text{OPT}}$ \\
        \midrule
        \multicolumn{2}{l}{\textit{Relation Extraction}} \\
        \textbf{Input} & \verb|[I]| Mirror Mirror, please help me extract the entity relationship triplet. \verb|[LR]| break up \verb|[TL]| The drama surrounding the high-profile divorce between Hollywood actors Johnny Depp and Amber Heard appears to be over as the couple reportedly reached an amicable settlement. \\
        \textbf{Output} & (\verb|[LR]|$_{\text{break up}}$, \texttt{Amber Heard}, \texttt{Johnny Depp}) \\
        \bottomrule
    \end{tabular}
    \caption{
        Case results obtained by the pretrained Mirror w/ Inst.
        The name of our proposed Mirror is borrowed from the magic mirror in \textit{Snow White and the Seven Dwarfs}.
        We hope to build a universal model that can help more people solve more problems.
    }
    \label{tab:case-study}
\end{table*}

\begin{figure*}[th]
    \centering
    \includegraphics[width=\textwidth]{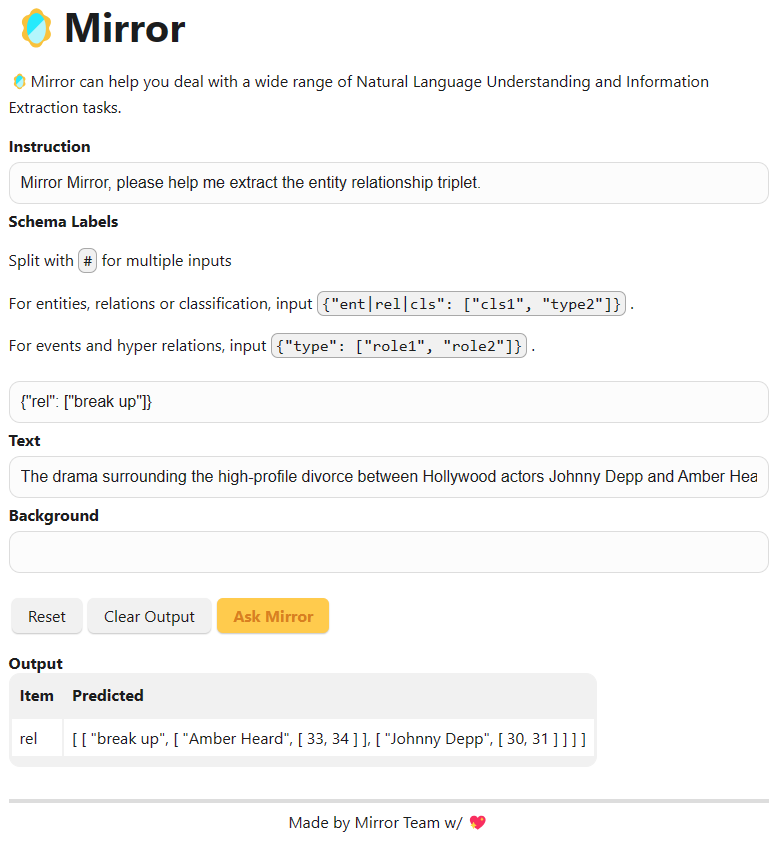}
    \caption{
        Mirror toolkit demonstration.
        The predicted relation label is converted to label string \textit{break up}.
        The positions shown in the predicted results are counted by tokens, so they do not match the input string characters.
        You can find the pretrained model weights and demo code in the repository and deploy it on your own machine: \url{https://github.com/Spico197/Mirror}
    }
    \label{fig:toolkit-demo}
\end{figure*}

\end{document}